\documentclass[letterpaper, 10 pt, conference]{ieeeconf}  

\IEEEoverridecommandlockouts                              
\usepackage[utf8]{inputenc}
\usepackage[T1]{fontenc}
\usepackage{graphicx}
\usepackage{caption}
\usepackage{caption}
\usepackage{tikz}
\usetikzlibrary{intersections}
\usepackage{pgfplots}
\pgfplotsset{compat=1.18}
\usepgfplotslibrary{fillbetween}

\usepackage{multirow}
\usepackage{booktabs}
\usepackage{array}
\usepackage{amsmath} 
\usepackage{amssymb}  
\usepackage{cuted}

\title{\LARGE \bf
Hazard-Aware Traffic Scene Graph Generation
}

\author{Yaoqi Huang, Julie Stephany Berrio, Mao Shan, and Stewart Worrall
\thanks{All authors are with Australian Center for Robotics (ACFR), The University of Sydney, NSW, Australia.
       {\tt\small y.huang, j.berrio, m.shan, s.worrall\}@acfr.usyd.edu.au.}}%
}

\begin{document}

\maketitle
\thispagestyle{empty}
\pagestyle{empty}

\begin{strip}
\vspace{-60px}
\centering

\begin{center}
    \centering
    \includegraphics[width=1\linewidth]{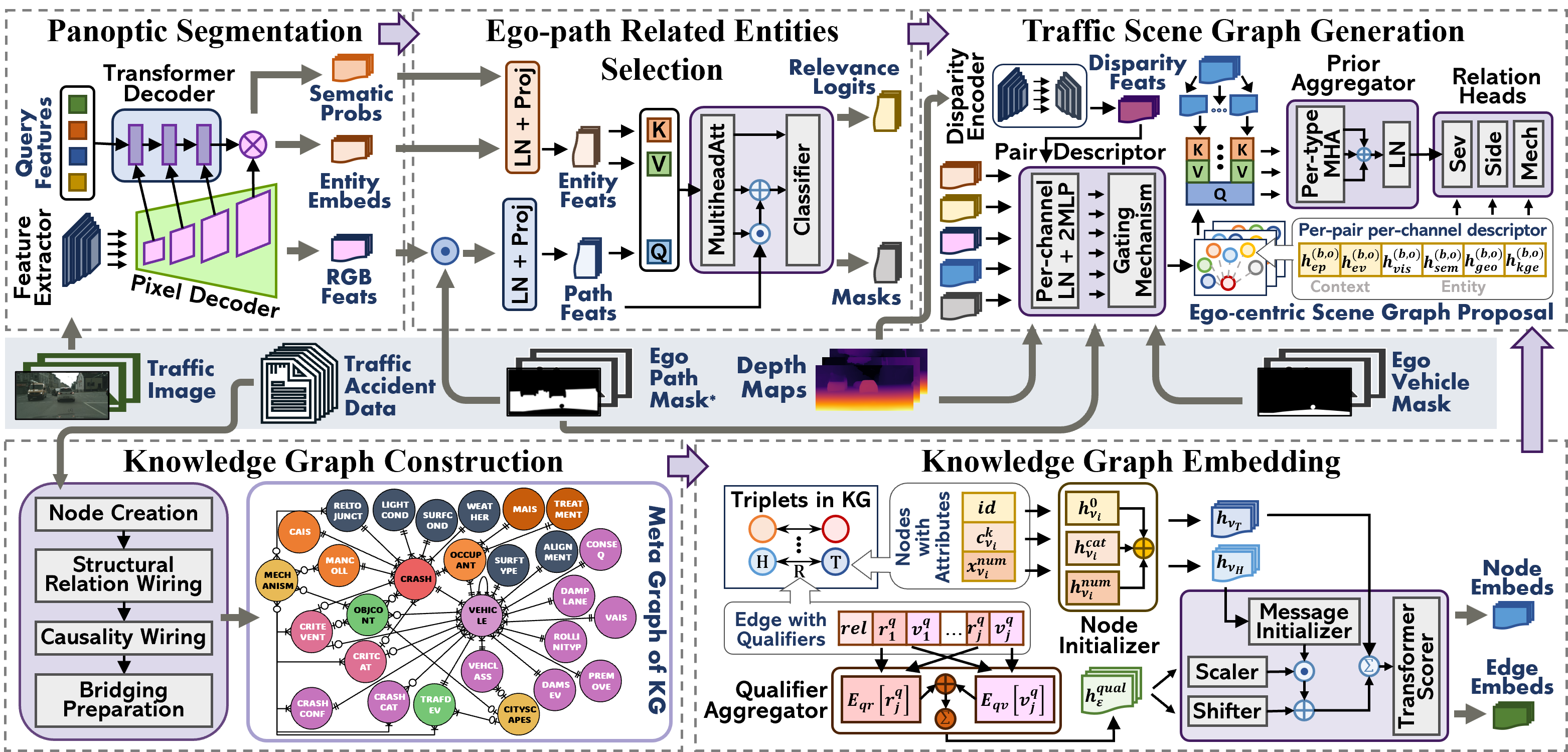}
    \captionof{figure}{\small Overview of our \underline{H}azard-\underline{A}ware \underline{T}raffic \underline{S}cene Graph Generation ({\small HATS}) model. The main scene graph branch (top) comprises three modules: 1) a Panoptic Segmentation (PS) Module for holistic perception of the surrounding environment, 2) an Ego-path Related Entities Selection (ERES) module that identifies and selects relevant candidate entities, and 3) a Traffic Scene Graph Generation (TSGG) module that computes ego-centric, traffic-specific relations among prominent hazards. The auxiliary knowledge branch (bottom) provides supplementary features and historical context to support entity representation and severity prediction within the TSGG module.}
    \label{fig:architecture}
\end{center}

\end{strip}

\begin{abstract}

Maintaining situational awareness in complex driving scenarios is challenging. It requires continuously prioritizing attention among extensive scene entities and understanding how prominent hazards might affect the ego vehicle. While existing studies excel at detecting specific semantic categories and visually salient regions, they lack the ability to assess safety-relevance. Meanwhile, the generic spatial predicates either for foreground objects only or for all scene entities modeled by existing scene graphs are inadequate for driving scenarios. To bridge this gap, we introduce a novel task, Traffic Scene Graph Generation, which captures traffic-specific relations between prominent hazards and the ego vehicle. We propose a novel framework that explicitly uses traffic accident data and depth cues to supplement visual features and semantic information for reasoning. The output traffic scene graphs provide intuitive guidelines that stress prominent hazards by color-coding their severity and notating their effect mechanism and relative location to the ego vehicle. We create relational annotations on Cityscapes dataset and evaluate our model on 10 tasks from 5 perspectives. The results in comparative experiments and ablation studies demonstrate our capacity in ego-centric reasoning for hazard-aware traffic scene understanding. 

\end{abstract}

\section{INTRODUCTION}

Despite statistics confirming distracted driving as one of the causes of traffic injuries and deaths \cite{2022_Choudhary}, it remains underappreciated \cite{who_mobile_2011}. 
In this context, helping drivers identify what truly deserves their attention becomes critical. Such guidance must raise situational awareness without distracting from the core driving maneuvers \cite{9758639}. Beyond accurate environment perception \cite{Fagnant}, this requires precise identification of hazards and instantly interpretable output. To address these demands, we propose \underline{H}azard-\underline{A}ware \underline{T}raffic \underline{S}cene Graph Generation ({\small HATS}), a model capable of (1) analyzing the entire traffic scenario, (2) identifying prominent hazards that warrant priority attention, (3) recognizing their traffic-specific relations to the ego vehicle, and (4) generating intuitive hazard-aware Traffic Scene Graphs ({\small TSGs}) to guide attention allocation.
Rather than limiting recognition to specific semantic classes, such as traffic lights \cite{s2tld}, 
our model aims to provide a comprehensive interpretation of the entire traffic scene. The Panoptic Segmentation ({\small PS}) module first localizes and classifies all entities as non-overlapping instances \cite{PanopticSegmentation}, enabling holistic scene perception. The Traffic Scene Graph Generation ({\small TSGG}) module then links these discrete entities to the ego vehicle, forming a systematic scene depiction.
Our model identifies prominent hazards that deserve priority attention \cite{CFHP} through two successive screenings. First, the Ego-path Related Entities Selection ({\small ERES}) module narrows candidates to those that could impact the ego path, including foreground road users (e.g. pedestrians with right-of-way) and background infrastructure (e.g. traffic devices), extending beyond the scope of dangerous road users \cite{9423525} or road obstacles. The resulting proposals are then further distilled to stress ego-centric relations with prominent hazards.

To ensure the utility of {\small TSG}s, we compute traffic-specific relations rather than generic visual-based ones such as `on' or `in' \cite{VisualGenome}. In this study, we explore accidents-aware relations. As illustrated in Fig.~\ref{fig:architecture}, the Knowledge Graph ({\small KG}) branch first structures isolated traffic accident data from public authorities \cite{NHTSA1,NHTSA2} into an integrated {\small KG} via a four-stage pipeline, followed by triplet attribute-aware KG embedding ({\small KGE}). The Scene Graph ({\small SG}) branch then jointly analyzes visual features, geometry cues, semantic information, and prior knowledge to reason the relations between each candidate entity and the ego vehicle in three perspectives (mechanism, side, and severity), producing triplets such as $\langle sidewipe, right, caution \rangle$.
As shown in Fig.~\ref{fig:viz_TSG}, in the final {\small TSG}, entities are color-coded across five severity levels to emphasize prominent hazards and suppress irrelevant ones. Each prominent hazard is annotated with a tag indicating its predicted relative side and the interaction mechanism with respect to the ego vehicle.

We summarize our key contributions as follows.
\begin{itemize}
    \item We propose {\small TSGG}, a new ego-centric problem formulation emphasizing prominent hazards and their traffic-specific relations, supported by relational annotations on Cityscapes as a baseline for future work.
    \item We propose a four-stage pipeline structuring isolated traffic accident data into a unified {\small KG}, paired with a {\small KGE} method that handles hyper-relational edges and multi-attributed nodes, outperforming baselines on three tasks.
    \item To our knowledge, it is the first work to leverage traffic accident data for traffic image interpretation, explicitly incorporating {\small KGE} as a supplementary modality and guiding severity assessment with historical accident records. 
\end{itemize}

\section{Related work}
\label{section:related_work}
\subsection{Graph-based Traffic Scene Understanding}
Scene graph generation ({\small SGG}) offers an explicitly structured representation with nodes as entities and edges as relations \cite{SGG}. However, general-purpose {\small SG}s are not capable of describing traffic scenes. 
Due to numerous traffic participants, the loosely enclosing bounding boxes tend to overlap heavily. The long-tail relation distribution pushes models to favor simple, frequent relations such as `on'. Although this may produce acceptable $R$@K \cite{lu2016visual}, it undermines the practicality and interpretability of the graph \cite{8954048}, particularly in traffic scenes. 
Panoptic Scene Graph ({\small PSG}) \cite{yang2022psg} extends scene graphs to include background-foreground relations via panoptic segmentation, but its application to traffic scenarios risks further obscuring critical triplets due to the increased number of entities and more intricate relations.

The key challenge is to represent the relationships between entities \cite{Reltr}. 
The distance, direction and lane-vehicle containment relations for vehicles are explored in \cite{9423525} using rule-based methods. Behavioral relationships among vehicles, pedestrians, and obstacles are studied in \cite{9575491}, such as ‘waiting for’. The behavior recognition study \cite{9197057} implicitly modeled interactions between the ego vehicle and specific entities, without explicitly analyzing or categorizing specific relations. Novelly, \cite{RS10K} visualized the corresponding relations between roads and components in traffic signs, helping drivers to grasp upcoming road information and restrictions.

Meanwhile, few benchmark datasets are designed for {\small TSG}. {\small HDD} dataset \cite{HDD} has real and artificial traffic scenes, annotating entities in nine categories with bounding boxes and their spatial or behavioral relations to the ego vehicle. {\small PSG} dataset \cite{yang2022psg} offers panoptic segmentation labels and involves a few traffic images with spatial or behavioral relations, but these samples are insufficient for traffic-specific learning. 

\subsection{Prior Knowledge and Knowledge Embedding}
{\small SGG} studies have supplemented visual features with external common sense knowledge for better relation reasoning, including statistical correlations \cite{8954048}, lexical databases such as WordNet \cite{WORDNET}, and well-structured semantic networks such as ConceptNet \cite{ConceptNet}. While common sense is undoubtedly an invaluable source of prior knowledge, specific domain knowledge might be more crucial if the application context shifts from general natural images to traffic scenarios. Meanwhile, existing studies directly analyze traffic data to draw conclusions about traffic accidents, whether from surveillance and dashcam videos \cite{you2020CTA, 8578469}, intersection traffic data \cite{9424477}, or social media \cite{9723740}. However, explicitly structuring traffic data as prior knowledge to facilitate understanding of traffic scene in other modality remains an unexplored area.

{\small KGs} offer a principled organization of such domain knowledge. {\small KGE} methods further map entities and relations into low-dimensional vector spaces to capture relational patterns. While seminal Graph Neural Networks ({\small GNN}) excel in general graph representation learning, they lack explicit relation embedding modeling and need modifications to support multi-relational {\small KG}s \cite{StarE}. Real-world {\small KG}s often contain hyper-relational facts beyond standard binary triplets. {\small StarE} \cite{StarE} addresses this with qualifier-aware message passing that jointly aggregates qualifiers alongside the main triplet while preserving their semantic roles. However, vanilla {\small StarE} \cite{StarE} offers no native support for arbitrary per-node properties, limiting its applicability in feature-critical settings.

\section{methodology}

\subsection{Ego-path Related Entities Selection}

Our {\small PS} module adopts a ResNet50-based Mask2Former \cite{mask2former}. While it comprehensively segments all entities, many are irrelevant to the immediate navigation of the ego vehicle. Computing triplets over all of them would be computationally expensive and could introduce noise that degrades the efficacy of {\small TSG}s. We therefore introduce the path-aware {\small ERES} module. It filters candidates to those relevant to the ego path and models the interaction between the ego path and traffic entities through a learnable cross-attention mechanism.

\textbf{(I)} We extract a path token from the shared mask-feature map $\mathbf{F}_\text{rgb} \in \mathbb{R}^{B \times C_\text{rgb} \times H_m \times W_m}$ using the resized binary path mask $\mathbf{M}_p \in \mathbb{R}^{B \times 1 \times H_m \times W_m}$. We then project it with a learnable mapping $\phi_\text{p}^\text{rgb}$ to obtain the path representation $\mathbf{F}_p$.
\begin{equation}
\mathrm{Pool}(\mathbf{F},\mathbf{M}) = \frac{\sum_{h,w} {\mathbf{F}}_{:,:,h,w} \cdot {\mathbf{M}}_{:,:,h,w}}{\sum_{h,w} {\mathbf{M}}_{:,:,h,w} + \epsilon}
\label{eq:masked_mean_pooling}
\end{equation}
\begin{equation}
\mathbf{F}_p = \phi_{p}^\text{rgb}\big(\mathrm{Pool}(\mathbf{F}_\text{rgb},\mathbf{M}_p)\big)  \in \mathbb{R}^{B \times 1 \times D}
\label{eq:path_rgb}
\end{equation}

\textbf{(II)} At its core, this module employs a multi-head cross-attention mechanism to quantify the relevance of each entity to the ego path, with the path representation serving as queries. The entity embeddings $\mathbf{E}_e \in \mathbb{R}^{B \times O \times D}$, derived from Mask2Former decoder and projected through a linear transformation, provide keys and values. The attention weights produced $\mathbf{A}_w \in \mathbb{R}^{B \times O}$ indicate the spatial and semantic relevance of each entity to the ego path. The path-conditioned entity summary $\mathbf{A}_e \in \mathbb{R}^{B \times D}$ represents `what other entities the path cares about in general'.
\begin{equation}
\mathbf{A}_w, \mathbf{A}_e = \mathrm{MHA}(\mathbf{F}_p, \mathbf{E}_e, \mathbf{E}_e)
\end{equation}

\textbf{(III)} For each entity $o$, we compute fused features $\mathbf{F}_{\text{fused}}^{(b,o)}$ by combining three complementary signals, where $\gamma_{\text{ctx}}$ and $\gamma_{\text{gate}}$ are learnable parameters. Finally, a two-layer {\small MLP} classifier predicts relevance logits $\mathbf{R} \in \mathbb{R}^{B \times O \times 2}$ that are used to select relevant queries for subsequent {\small SGG} module.
\begin{equation}
\mathbf{F}_{\text{fused}}^{(b,o)} = \mathbf{E}_e^{(b,o)} + \gamma_{\text{ctx}} \cdot \mathbf{A}_e^{(b)} + \gamma_{\text{gate}} \cdot \mathbf{A}_w^{(b,o)} \cdot \mathbf{F}_p^{(b)}
\label{eq:fused_feature}
\end{equation}
\begin{equation}
\mathbf{R}^{(b,o)} = \mathbf{W}_2 \cdot \phi_{\text{GELU}}(\mathbf{W}_1 [\mathbf{F}_{\text{fused}}^{(b,o)}; \mathbf{A}_w^{(b,o)}] + \mathbf{b}_1) + \mathbf{b}_2
\label{eq:relevance_classifier}
\end{equation} 
\begin{equation}
\Pi^{(b)}=\left\{\,o\ \middle|\ \arg\max_{c\in\{0,1\}}\mathbf{R}^{(b,o)}_{[c]}=1 \right\}
\label{eq:keep_mask}
\end{equation}

\subsection{Knowledge Graph Construction}

The source traffic accident data used to build our {\small KG} is derived from field crash investigation files \cite{NHTSA1,NHTSA2} distributed by the (US Department of Transportation, National Highway Traffic Safety Administration ({\small NHTSA}). As indicated in Table \ref{table:KG Components}, our {\small KG} is built through a four-stage pipeline, with strategies to ensure per-stage and overall quality.

\begin{table*}[th!]
\centering
\setlength{\tabcolsep}{3pt}
\renewcommand{\arraystretch}{0.8}  
\scriptsize
\caption{\small Core component explanations and statistics of the Traffic Accident Knowledge Graph (side-by-side node/edge view)}
\label{table:KG Components}
\begin{tabular}{%
  >{\centering\arraybackslash}m{0.35cm}  
  >{\centering\arraybackslash}m{2.0cm}  
  >{\raggedright\arraybackslash}m{1.9cm}
  >{\raggedright\arraybackslash}m{1.9cm}
  >{\centering\arraybackslash}m{0.35cm}  
  >{\centering\arraybackslash}m{0.9cm}  
  >{\raggedright\arraybackslash}m{2.65cm}
  >{\raggedright\arraybackslash}m{3.2cm}
  >{\centering\arraybackslash}m{3.0cm}  
}
\toprule
\multicolumn{4}{c}{\textbf{Node section}} & \multicolumn{5}{c}{\textbf{Edge section}} \\ [-1pt]
\cmidrule(r){1-4}\cmidrule(l){5-9}
\textbf{Stg} & \textbf{Class} & \multicolumn{2}{c}{\textbf{Type(count)}} &
\textbf{Stg} & \textbf{Class} & \multicolumn{2}{c}{\textbf{Type(count)}} & \textbf{Head$\rightarrow$Tail} \\ [-1pt]
\cmidrule(r){1-4}\cmidrule(l){5-9}

\multirow{25}{*}{I} & 
\multirow{4}{*}{\shortstack{Environment \\ Condition}} &
\shortstack[l]{LIGHTCOND(6)\\ RELTOJUNCT(5)}
& 
\shortstack[l]{WEATHER(12)\\ SURFCOND(9)}
&
\multirow{18}{*}{II} & \multirow{18}{*}{\shortstack{Schema- \\ driven \\ Relation}} &
\shortstack[l]{LightCondition(3324)\\ RelationToJunction(3340)}
&
\shortstack[l]{WeatherCondition(3322)\\ SurfaceCondition(3326)}
& 
\shortstack[c]{Crash $\rightarrow$ \\ Environment Condition}
\\[-1pt] 
\cmidrule(lr){3-4}\cmidrule(lr){7-9}
 & &     
\shortstack[l]{SURFTYPE(6)}
& 
\shortstack[l]{ALIGNMENT(4)}
& & &
\shortstack[l]{SurfaceType(5764)}
& 
\shortstack[l]{RoadwayAlignment(5758)}
& 
\shortstack[c]{Vehicle$\rightarrow$Env. Condition}
\\[-1pt] 
\cmidrule(lr){2-4}\cmidrule(lr){7-9}
&
\shortstack[c]{Crash Attributes}
&
\shortstack[l]{MANCOLL(7)}
&
\shortstack[l]{CAIS(9)}
& & &
\shortstack[l]{CollisionManner(3272)}
&
\shortstack[l]{MostSevereInjuryInCrash(3331)}
&
\shortstack[c]{Crash$\rightarrow$Crash Attribute}
\\[-1pt] 
\cmidrule(lr){2-4}\cmidrule(lr){7-9}

& \shortstack[c]{Vehicle \\ Attributes} &
\shortstack[l]{VEHCLASS(29)\\ CRASHCAT(6)\\ CRASHCONF(10) \\ CONSEQ(8)}
&
\shortstack[l]{DAMPLANE(9)\\ DAMSEV(4)\\ VAIS(10)\\ PREMOVE(20) \\ ROLLINITYP(11)}
& & &
\shortstack[l]{InstanceOf(5935)\\ CrashCategory(5765)\\ CrashConfiguration(5765) \\ GeneralConsequence(567)}
&
\shortstack[l]{VehicleDamagePosition(5862)\\ DamageSeverityLevel(4513)\\ MostSevereInjuryInVehicle(5935)\\ Pre-crashVehicleMovement(5756) \\ Post-crashRolloverType(5499)}
&
\shortstack[c]{Vehicle$\rightarrow$\\ Vehicle Attribute}
\\[-1pt] 
\cmidrule(lr){2-4}\cmidrule(lr){7-9}

& \shortstack[c]{Occupant Attributes} &
\shortstack[l]{TREATMENT(10)}
&
\shortstack[l]{MAIS(9)}
& & &
\shortstack[l]{TreatmentReceived(5782)}
&
\shortstack[l]{MostSevereInjury(6487)}
&
\shortstack[c]{Occupant$\rightarrow$Occupant Attr.}
\\[-1pt] 
\cmidrule(lr){2-4}\cmidrule(lr){7-9}

& Crash &
\multicolumn{2}{c}{\shortstack{CRASH(3331)(in 7 manners)}}
& & &
\shortstack[l]{VehicleInvolved(5935)\\ OccupantInvolved(6487)}
&
\shortstack[l]{EntityInvolved(2476)}
&
\shortstack[c]{Crash$\rightarrow$Passenger \\ /Driver/Entity/Vehicle}
\\[-1pt] 
\cmidrule(lr){2-4}\cmidrule(lr){7-9}

& Vehicle &
\multicolumn{2}{c}{\shortstack{VEHICLE(5935)(in 29 classes)}}
& & &
\multicolumn{2}{c}{\shortstack[l]{ContactWith(3294)}}
& 
\shortstack[c]{Vehicle$\rightarrow$Vehicle}
\\[-1pt] 
\cmidrule(lr){2-4}\cmidrule(lr){7-9}

& Occupant &
\multicolumn{2}{c}{\shortstack{OCCUPANT(6487)(Driver/Passenger)}}
& & &
\multicolumn{2}{c}{\shortstack[l]{HasOccupant(6487)}}
&
\shortstack[c]{Vehicle$\rightarrow$Occupant}
\\[-1pt] 
\cmidrule(lr){2-4}\cmidrule(lr){7-9}

& \shortstack{Scene Entity} &
\shortstack[l]{OBJCONT(35)} & \shortstack[l]{TRAFDEV(9)}
& & &
\shortstack[l]{Contactwith(2517)} & \shortstack[l]{SignBestControlsTraffic(5764)}
& \shortstack[c]{Vehicle$\rightarrow$Entity/Device}
\\[-1pt] 
\cmidrule(lr){2-4}\cmidrule(lr){5-9}

& \multirow{3}{*}{Cause} &
\multirow{3}{*}{\shortstack[l]{CRITEVENT(50)}} &
\multirow{3}{*}{\shortstack[l]{CRITCAT(8)}}
& \multirow{3}{*}{III} & \multirow{3}{*}{Causality} &
\shortstack[l]{LeadTo(9141)} & \shortstack[l]{MakeCrashImminentFor(11891)}
& \shortstack[c]{Vehicle/Entity/Event/Factor\\$\rightarrow$Crash/Vehicle/Entity}
\\[-1pt] 
\cmidrule(lr){7-9}

& & & & & &
\multicolumn{2}{c}{\shortstack[l]{ImplicatedBy (actor reports)(5163) / (victim reports)(4909)}}
&
\shortstack[c]{Vehicle/Entity$\rightarrow$Event/Factor}
\\[-1pt] 
\cmidrule(lr){1-4}\cmidrule(lr){5-9}

IV & Bridging Node &
\shortstack[l]{CITYSCAPES(19)} &
\shortstack[l]{MECHANISM(8)}
& IV & Bridging &
\multicolumn{2}{c}{\shortstack[l]{CorrespondsTo (cityscapes)(57) / (mechanisms)(64)}}
& \shortstack[c]{Stg-I Node$\rightarrow$Bridging Node}
\\[-1pt] 

\bottomrule
\end{tabular}
\end{table*}

\textbf{(I)} 
We construct nodes from the complete yearly NHTSA dataset, which comprises 39 CSV files documenting various crash features. We retain variables that are visually identifiable in traffic scene images or serve as essential descriptors of scene characteristics
An extraction layer codified as a contract governs inner-node structure (e.g., name, label, type, and properties) and inter-node hierarchy (e.g., entity and attribute), aligning heterogeneous data and enabling cross-table mapping. A decoding layer then normalizes and cleans the data prior to graph ingestion to prevent duplicates and ensure idempotence, ultimately producing 16,039 nodes across 26 types instantiating four-level entities and five-level attributes.

\textbf{(II)} The Schema-driven structural relation wiring stage is designed to materialize the deterministic relations prescribed by the table schema. The structural relations explicitly encoded in Stage I, including attribution, consequence, membership, contact, and taxonomy, are decoded here into 122,263 edges across 25 types, connecting the previously created nodes. The coherence between nodes and edges (e.g. $Vehicle$ and $Vehicle\_Involved$) and between edges (e.g. $Occupant\_Involved$ and $Has\_Occupant$), as well as the consistency between the {\small KG} and the source tables, are automatically verified and followed by targeted manual unit tests in the edges. Full compliance in these examinations further confirms the validity of Stage I.

\textbf{(III)} Beyond explicit structural relations, we enrich the {\small KG} with the causality underlying each crash, introducing 31,104 causal edges in three types. Edge properties such as $Implicated\_By (source=actor|victim, role=event|factor,witness)$ preserve provenance, explicitly revealing the actor at-fault, the exonerated participants, and the critical factors and events that led to each crash.

\textbf{(IV)} We introduce bridge nodes and mapping edges to align the {\small KG} with external taxonomies, linking existing nodes to semantic categories, and mapping existing edges to our designed graph predicate schema. These cross-ontology alignment hooks promote homogeneity between the {\small KG}, upstream segmentation, and downstream scene graph structure.

Ultimately, we elevated multiple isolated flat tables into our integrated traffic accident {\small KG}.  
It contains 16,066 nodes and 153,488 edges, aligned across ontologies. We then export our heterogeneous multi-relational $\mathcal{KG} = (\mathcal{N}_\text{kg}, \mathcal{E}_\text{kg}, \mathcal{T}_\text{kg})$ in triplet format $\tau=(\nu_\text{head},\varepsilon,\nu_\text{tail})$, where $\tau \in \mathcal{T}_\text{kg}$, $\nu_\text{head},\nu_\text{tail} \in \mathcal{N}_\text{kg}$ and $\varepsilon \in \mathcal{E}_\text{kg}$. Compared with other common-sense {\small KG}s, our {\small KG} provides extensive domain-specific knowledge and negative facts for subsequent {\small KGE} and {\small SGG}.

\subsection{Knowledge Graph Embedding}
\label{subsection:KGE}
Our KGE module uses rich node and edge properties to encode triplets rather than conventional ID-based embeddings~\cite{TransR}. The relation-aware message is adaptively modulated by qualifier-derived parameters. A transformer-based triplet scorer further enables attention-driven interaction.

\textbf{(I)} 
Node embeddings are initialized by fusing all respective node properties. We first create embeddings for seven categorical properties: three universal ones for all node types (name, type, and level), and four type-specific ones (collision manner for {\small \textit{CRASH}} nodes, class for {\small \textit{VEHICLE}} nodes, and most severe injury and treatment for {\small \textit{OCCUPANT}} nodes).
For property $k$, we define an embedding matrix $E_\text{cat}^k \in \mathbb{R}^{|C_k| \times d}$, where $|C_k|$ is the vocabulary size of property $k$ and $d$ is the dimension of the model. For node $\nu_i$ with $K$ categorical properties ($K \leq 7$), its categorical value of property $k$ is $c_{\nu_i}^k$, with its overall categorical embedding denoted below.
\begin{equation}
\mathbf{h}_{\nu_i}^\text{cat} = \sum_{k=1}^K\mathbf{E}_\text{cat}^k[c_{\nu_i}^k]
\end{equation}
In particular, we embed five numerical properties for {\small \textit{CRASH}} nodes, including vehicle count, event count, and three crash time features. As stated in \eqref{crashembed}, the embedding $\mathbf{x}_{\nu_i}^\text{num} \in \mathbb{R}^{5}$ is encoded into $d$ dimensions using a 2-layer {\small MLP} with {\small ReLU}, where $\mathbf{W}_3 \in \mathbb{R}^{d \times 5}$, $\mathbf{W}_4 \in \mathbb{R}^{d \times d}$, and $\mathbf{b}_3, \mathbf{b}_4 \in \mathbb{R}^{d}$.

\begin{equation}
\mathbf{h}_{\nu_i}^\text{num} = \mathbf{W}_4 \cdot \phi_\mathrm{ReLU}(\mathbf{W}_3 \mathbf{x}_{\nu_i}^\text{num} + \mathbf{b}_3) + \mathbf{b}_4
\label{crashembed}
\end{equation}
We then obtain literal embeddings $\mathbf{h}_i^{lit} \in \mathbb{R}^{d}$ for {\small \textit{CRASH}} nodes by combining their categorical and numerical embeddings, while other nodes use categorical embeddings directly.

\begin{equation}
\mathbf{h}_{\nu_i}^\text{lit} = \begin{cases}
\mathbf{h}_{\nu_i}^\text{cat} + \mathbf{h}_{\nu_i}^\text{num}, & \text{if both exist} \\
\mathbf{h}_{\nu_i}^\text{cat}, & \text{else}
\end{cases}
\end{equation}
Finally, we fuse $\mathbf{h}_{\nu_i}^\text{lit}$ with the node {\small ID} embedding $\mathbf{h}_{\nu_i}^{(0)}$ through a linear mixer, where $\mathbf{W}_\text{lit} \in \mathbb{R}^{d \times d}$ and $\mathbf{b}_\text{lit} \in \mathbb{R}^{d}$.
\begin{equation}
\mathbf{h}_{\nu_i} = \mathbf{h}_{\nu_i}^{(0)} + \mathbf{W}_\text{lit} \mathbf{h}_{\nu_i}^\text{lit} + \mathbf{b}_\text{lit}
\end{equation}

\textbf{(II)} 
We embed edge properties of our Causality edges, Bridging egdes, and `Contact\_with' edge as `qualifier' \cite{StarE}. For edge $\varepsilon$, the set of qualifiers is $\mathcal{Q}_{\varepsilon} = \{(r_j^q,v_j^q)\}_{j=1}^{|\mathcal{Q}_{\varepsilon}|}$, where $r_j^q$ is the qualifier type and $v_j^q$ is the qualifier value, such as $(source: actor)$. First, we create an embedding matrix for the qualifier relations $\mathbf{E}_{qr} \in \mathbb{R}^{|R_q| \times d}$ and another one for the qualifier values $\mathbf{E}_{qv} \in \mathbb{R}^{|V_q| \times d}$. Then we aggregate embeddings $\mathbf{q}_j$ of all qualifiers associated with each edge to obtain a qualifier vector per-edge $\mathbf{h}_{\varepsilon}^\text{qual}$. For edges without properties, there is no qualifier and thus $\mathbf{h}_{\varepsilon}^\text{qual}=0$.

\begin{equation}
\mathbf{h}_{\varepsilon}^\text{qual} = \sum_{j=1}^{|\mathcal{Q}_{\varepsilon}|} \mathbf{q}_j; \; \mathbf{q}_j = \mathbf{E}_{qr}[r_j^q] + \mathbf{E}_{qv}[v_j^q]
\end{equation}

\textbf{(III)} 
We stack $L$ layers of relation-aware message passing to iteratively refine node representations. 
In layer $\ell \in \{0,\dots,L-1\}$, we update each node using messages from its one-hop neighbors, so that each node embedding $\mathbf{h}_{\nu}^{L}$ can aggregate information from up to $L$-hop neighborhoods. 
Edge qualifiers act as modulators through feature-wise linear modulation ({\small FiLM)}~\cite{FiLM} to adaptively scale and shift messages. 
First, for each triplet $\tau=(\nu_\text{head},\varepsilon,\nu_\text{tail})$, we compute the initial message $\tilde{\mathbf{m}}_{\varepsilon}$ passed from $\nu_\text{head}$ using a relation-specific linear transformation, where $rel$ is the relation type for $\varepsilon$, $\mathbf{W}_\text{rel} \in \mathbb{R}^{d \times d}$, and $\mathbf{b}_\text{rel} \in \mathbb{R}^{d}$. 
The message is then modulated using the qualifier vector $\mathbf{h}_{\varepsilon}^\text{qual}$.
\begin{equation}
\tilde{\mathbf{m}}_{\varepsilon} = \mathbf{W}_\text{rel} \mathbf{h}_{\nu_\text{head}}^\ell + \mathbf{b}_\text{rel}
\end{equation}
\begin{equation}
\boldsymbol{\gamma}_{\varepsilon} = \mathbf{W}_{\gamma} \mathbf{h}_{\varepsilon}^\text{qual} + \mathbf{b}_{\gamma}; \;
\boldsymbol{\beta}_{\varepsilon} = \mathbf{W}_{\beta} \mathbf{h}_{\varepsilon}^\text{qual} + \mathbf{b}_{\beta}
\end{equation}
\begin{equation}
\mathbf{m}_{\varepsilon} = \left(1+\boldsymbol{\gamma}_{\varepsilon}\right) \odot \tilde{\mathbf{m}}_{\varepsilon} + \boldsymbol{\beta}_{\varepsilon}
\end{equation}
Finally, the messages are aggregated at each destination node $\nu_\text{tail}$, where $\mathcal{E}(\nu_\text{tail})$ denotes the set of its incoming edges.
\begin{equation}
\mathbf{h}_{\nu_\text{tail}}^{\ell+1} = \mathrm{LN} \left( \mathbf{h}_{\nu_\text{tail}}^{\ell} + \sum_{\varepsilon \in \mathcal{E}(\nu_\text{tail})} \mathbf{m}_{\varepsilon} \right)
\end{equation}

\textbf{(IV)} 
After $L$ layers of message passing, we stack the head node embedding $\mathbf{h}_{\nu_\text{head}}^L$ and a relation token $\mathbf{r}_{\varepsilon}$, where $\mathbf{E}_\text{rel} \in \mathbb{R}^{|N_r| \times d}$ has one $d$-dimensional row per relation type. 
We feed this 2-token sequence through a Transformer encoder to produce the query representation $\mathbf{z}$. 
The triplet score is computed by comparing $\mathbf{z}$ with all tail embeddings $\mathbf{h}_{\nu_\text{tail}}^L$.
\begin{equation}
\mathbf{r}_{\varepsilon} = \mathbf{E}_\text{rel}[rel] + \mathbf{W}_\text{seq} \mathbf{h}_{\varepsilon}^\text{qual} + \mathbf{b}_\text{seq}
\end{equation}
\begin{equation}
\mathbf{z} = \mathrm{TransformerEncoder}([\mathbf{h}_{\nu_\text{head}}^L; \mathbf{r}_{\varepsilon}])_1 \in \mathbb{R}^{d}
\end{equation}
\begin{equation}
s_\text{kge}(\tau) = \mathbf{z}^{\top}\mathbf{h}_{\nu_\text{tail}}^L
\end{equation}

\textbf{(V)} 
Since each head-edge pair may have multiple positive tail nodes and each head-tail pair may admit several positive relations, we adopt a 1-to-N scoring strategy with filtered multi-label classification to train our model.
For each head-edge query $\left(\nu_\text{head},\varepsilon\right)$, we construct a smoothed multi-hot target vector $\mathbf{y} \in \mathbb{R}^{|\mathcal{N}_\text{kg}|}$, where $y_i=1-\theta$ if $\tau=(\nu_\text{head},\;\varepsilon,\;\nu_i)$ is a positive triplet; otherwise $y_i=\theta$. We score all candidate triplets and compute the binary cross-entropy loss $\mathcal{L}_{tail}$ for the tail prediction. Then, we define a reciprocal relation $rel^{-1}$ for each $rel$ and use the same procedure to compute the loss $\mathcal{L}_\text{head}$ of head prediction for each $\left(\nu_\text{tail},\varepsilon'\right)$ query. The final objective is to minimize the loss that combines both directions.
\begin{equation}
\mathcal{L} = \frac{1}{2}(\mathcal{L}_\text{tail} + \mathcal{L}_\text{head})
\end{equation}

\subsection{Traffic Scene Graph Generation}

The {\small TSGG} module produces ego-centric graphs by integrating visual features, 3D structural cues, semantic information, and prior knowledge. A lightweight disparity encoder complements the {\small RGB} backbone. A gated fusion strategy combines these multiple cues into a robust ego–entity pair descriptor. Dedicated relation heads then infer the effect mechanism, relative side, and severity level for each relevant entity, yielding ego-centric relational triplets.

\textbf{(I)} 
To complement {\small RGB} features with geometry cues for downstream reasoning, we build a lightweight disparity encoder for single-channel disparity images. It mirrors the ResNet stride schedule with a single \(3\times3\) residual block per stage. A compact Feature Pyramid Network ({\small FPN}) aggregates these stages to produce a single fused disparity feature map $\mathbf{F}_\text{disp} \in \mathbb{R}^{B \times C_\text{disp} \times H \times W}$ at the resolution of the mask-feature.

\textbf{(II)}
Each candidate entity $o\in\Pi^{(b)}$ in each image $b$ is depicted using four complementary representations. The visual representation $\mathbf{h}^{(b,o)}_\text{vis}$ is obtained by projecting the concatenation of {\small RGB} and disparity features with $\phi_\text{vis}$. The two feature vectors are extracted from the mask-feature map $\mathbf{F}_\text{rgb}$ and the disparity-feature map $\mathbf{F}_\text{disp}$ within the entity mask $\mathbf{M}_e^{(b,o)}$. The semantic representation $\mathbf{h}_\text{sem}^{(b,o)}$ projects the concatenation of the decoder embedding $\mathbf{E}_e^{(b,o)}$ and the path-relevance logits $\mathbf{R}^{(b,o)}$ with $\phi_\text{sem}$. The geometry representation $\mathbf{h}_\text{geo}^{(b,o)}$ encodes the overlap score with the ego path and the proximity to the ego vehicle with the projection. The {\small KGE} representation $\mathbf{h}_\text{kge}^{(b,o)}$ is retrieved and projected from the {\small KG} embedding of the bridging node corresponding to the predicted entity semantic class. We then build representations $\mathbf{h}_\text{ep}$ and $\mathbf{h}_\text{ev}$ for the ego path and the ego vehicle in the same way as $\mathbf{h}_\text{vis}$ but using the corresponding mask $\mathbf{M}_p$ or $\mathbf{M}_v$. They are replicated across all entities in the same image. After getting the compact representation of each ego-entity pair $\mathbf{\hat{h}}^{(b,o)}_\text{pair}$, we obtain the final pair descriptor $\mathbf{h}_\text{pair} \in \mathbb{R}^{D_\text{pair}}$ using a learned gating mechanism. It allows the model to adaptively weight each cue per pair and per channel via $\mathbf{g}_\text{pair}^{(b,o)} \in (0,1)^{6D_\text{pair}}$ for the downstream heads.
\begin{equation}
\mathbf{\hat{h}}_\text{pair} = \big[\,\mathbf{h}_\text{ep} \;\Vert\; \mathbf{h}_\text{ev} \;\Vert\; \mathbf{h}_\text{vis} \;\Vert\; \mathbf{h}_\text{sem} \;\Vert\; \mathbf{h}_\text{geo} \;\Vert\; \mathbf{h}_\text{kge} \,\big]
\label{eq:ent_temp_rep}
\end{equation}
\begin{equation}
\mathbf{h}_{\text{pair}}
= \sum
\bigg(\sigma\Big(\mathbf{W}_6\phi_{\mathrm{GELU}}\big(\mathbf{W}_5\mathrm{LN}(\mathbf{\hat{h}}_\text{pair})\big)\Big)\bigg) \odot \mathbf{\hat{h}}_\text{pair}
\label{eq:ent_rep}
\end{equation}

\textbf{(III)}
We inject structured traffic accident priors using multi-head attention for severity prediction. We determine 7 severity-related {\small KG} node groups $\mathcal{N}^\text{kg}_\text{sev}$ that include {\small CAIS}, {\small VAIS}, {\small MAIS}, {\small DAMSEV}, {\small CONSEQ}, {\small TREATMENT}, and {\small ROLLINITYP}. The aligned pair query $\mathbf{h}^{'}_\text{pair}$ attends to the specific node embeddings $\mathbf{h}_{\nu_{tp}}$ in each group $tp$, producing a compact type-specific prior vector. All type priors are then concatenated and normalized for severity head.
\begin{equation}
\mathbf{h}^{(b,o)}_\text{prior}
=\mathrm{LN}\Big(\big\Vert_{{tp}\in\mathcal{N}^\text{kg}_\text{sev}}\phi_\text{prior}\big(\mathrm{MHA}(\mathbf{h}^{'(b,o)}_\text{pair},\mathbf{h}_{\nu_{tp}},\mathbf{h}_{\nu_{tp}})\big)\Big)
\label{eq:prior_agg_type}
\end{equation}

\textbf{(IV)}
The \textit{Mechanism Head} classifies the potential effect mechanism of each entity towards the ego vehicle $\mathbf{s}_\text{mech}^{(b,o)} \in \mathbb{R}^{8}$. The pair descriptor is aligned with the {\small KGE} space and matched with the 8 mechanism prototypes in {\small KGE} $\mathbf{h}_{\nu_\text{mech}}$ using cosine similarity with a learned temperature $tem$. The \textit{Side Head} predicts the relative position to the ego vehicle $s_\text{side}^{(b,o)} \in \mathbb{R}^{3}$. Similarly to \eqref{eq:ent_temp_rep} and \eqref{eq:ent_rep}, it first fuses the 5 image-based representations ($\mathbf{h}_\text{ep}$, $\mathbf{h}_\text{ev}$, $\mathbf{h}_\text{vis}$, $\mathbf{h}_\text{sem}$, and $\mathbf{h}_\text{geo}$) with an element-wise gating network and then classifies the fused vector $\mathbf{\hat{h}}_\text{side}$ with a 2-layer $\mathrm{MLP}$. The \textit{Severity Head} classifies each entity into 4 levels $\mathbf{s}_\text{sev}^{(b,o)} \in \mathbb{R}^{4}$ based on how seriously it deserves attention. It analyzes the concatenated vector $\mathbf{h}_{\text{sev}}$ of the pair descriptor and the prior vector $\mathbf{h}_{\text{prior}}$.
\begin{equation}
\mathbf{s}_\text{mech} = tem \cdot\mathrm{cos}\Big(\phi_\mathrm{ReLU}\big(\mathrm{LN}(\mathbf{W}_7\mathbf{h}_{\text{pair}}+\mathbf{b}_7)\big),\mathbf{h}_{\nu_\text{mech}}\Big) 
\end{equation}
\begin{equation}
\mathbf{s}_\text{side} = \mathbf{W}_9\phi_\mathrm{GELU}\big(\mathbf{W}_8 \mathrm{LN}(\mathbf{\hat{h}}_\text{side})+\mathbf{b}_8\big)+\mathbf{b}_9\
\end{equation}
\begin{equation}
\mathbf{s}_\text{sev} = \mathbf{W}_{11}\phi_\mathrm{GELU}\big(\mathrm{LN}(\mathbf{W}_{10} \mathbf{h}_{\text{sev}}+\mathbf{b}_{10})\big)+\mathbf{b}_{11}
\end{equation}

\section{Experiments and Results}

Due to the absence of suitable benchmark datasets for this task, we build upon the Cityscapes dataset \cite{Cityscapes}, leveraging its traffic images and panoptic segmentation labels. We create our own relation labels, describing the effect mechanism (control, edge\_proximity, fixed\_object\_near\_edge, head\_on, intersection, rear\_end, sideswipe, cross\_traffic\_conflict), relative side (left, front, right), and severity level (info, caution, imminent, relevant\_but\_not\_critical) of each relevant entity.

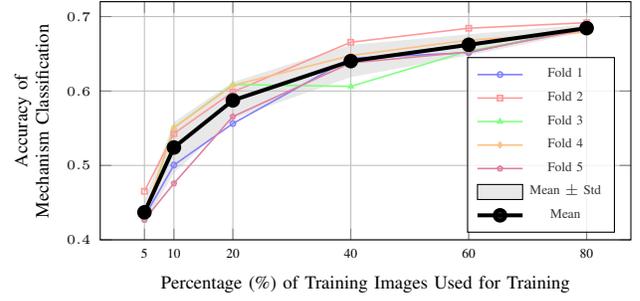
\begin{figure}[!t]
\centering
\begin{tikzpicture}
\begin{axis}[
    xlabel={Percentage (\%) of Training Images Used for Training},
    ylabel={Accuracy of \\ Mechanism Classification},
    ylabel style={align=center},
    legend pos=south east,
    legend style={font=\tiny, fill=white, fill opacity=0.8, draw opacity=1},
    grid=major,
    width=\columnwidth,
    height=0.55\columnwidth,
    xshift=-5pt,
    ymin=0.40,
    ymax=0.72,
    xtick={5,10,20,40,60,80},
    xticklabels={5,10,20,40,60,80},
    tick label style={font=\tiny},
    label style={font=\scriptsize},
]

\addplot[color=blue!50, mark=o, mark size=1pt, line width=0.6pt] coordinates {
    (5,0.430882944) (10,0.500529264) (20,0.556285977) (40,0.644106263) (60,0.65149541) (80,0.685834713)
};
\addlegendentry{Fold 1}

\addplot[color=red!40, mark=square, mark size=1pt, line width=0.6pt] coordinates {
    (5,0.465212399) (10,0.542856985) (20,0.598692695) (40,0.665511191) (60,0.68434629) (80,0.691779005)
};
\addlegendentry{Fold 2}

\addplot[color=green!50, mark=triangle, mark size=1pt, line width=0.6pt] coordinates {
    (5,0.428770799) (10,0.550704754) (20,0.608787097) (40,0.606188957) (60,0.653662473) (80,0.682823077)
};
\addlegendentry{Fold 3}

\addplot[color=orange!50, mark=diamond, mark size=1pt, line width=0.6pt] coordinates {
    (5,0.433017575) (10,0.550818588) (20,0.607630734) (40,0.647666426) (60,0.668348683) (80,0.679038074)
};
\addlegendentry{Fold 4}

\addplot[color=purple!50, mark=pentagon, mark size=1pt, line width=0.6pt] coordinates {
    (5,0.42690766) (10,0.475820411) (20,0.565673124) (40,0.637762194) (60,0.652477371) (80,0.682971712)
};
\addlegendentry{Fold 5}

\addplot[
    name path=upper,
    draw=none,
    forget plot,
] coordinates {
    (5,0.4529) (10,0.5583) (20,0.6121) (40,0.6619) (60,0.6763) (80,0.6892)
};

\addplot[
    name path=lower,
    draw=none,
    forget plot,
] coordinates {
    (5,0.4210) (10,0.4900) (20,0.5627) (40,0.6186) (60,0.6478) (80,0.6798)
};

\addplot[fill=black!30, fill opacity=0.3] fill between[of=upper and lower];
\addlegendentry{Mean $\pm$ Std}

\addplot[
    color=black,
    line width=1.5pt,
    mark=*,
    mark size=2pt,
] coordinates {
    (5,0.4370) (10,0.5241) (20,0.5874) (40,0.6402) (60,0.6621) (80,0.6845)
};
\addlegendentry{Mean}
\end{axis}
\end{tikzpicture} \\[-6pt]
\caption{ \small Inference performance vs. training set size (5\%–80\% of total training set). For each size, five models were trained with five-fold splits, with 20\% of training images held out for validation per fold.}
\label{fig:cv_label_curve}
\vspace{-3mm}
\end{figure}

Due to the high cost of annotation, we labeled 820 images for training and testing. Hyperparameters were selected using 5-fold cross-validation. As shown in Fig.~\ref{fig:cv_label_curve}, the model performance exhibits a strong positive correlation with the training set size, increasing from 0.437±0.016 at 5\% of total training images to 0.684±0.005 at 80\%. The decreasing standard deviation demonstrates that larger training sets produce more stable and generalizable models. The continued upward trend suggests that our model would benefit from additional training data beyond the currently available images. Additionally, the 8:1:1 ratio was used to split the 153,488 triplets in our {\small KG} to train, validate and test our {\small KGE} module. 

The experimental platform uses a single A40 {\small GPU}. Our {\small KG} branch uses the training strategy elaborated in Sec. \ref{subsection:KGE}, while {\small SG} branch uses complementary loss functions for three modules. The {\small PS} module adopts sigmoid cross-entropy mask loss, cross-entropy classification loss, and dice loss \cite{mask2former}. The {\small ERES} module and three {\small TSGG} heads are supervised by respective frequency-balanced softmax focal loss. A warm-up schedule is applied to establish a stable initialization. During the warm-up period, all {\small TSGG} heads are zeroed out, allowing the {\small ERES} module to receive focused gradient updates and the {\small PS} module to be jointly fine-tuned. After warming up, the supervisory focus shifts towards the {\small TSGG} heads, while still fine-tuning the {\small ERES} module at reduced weight, and the {\small PS} module is frozen. Optimization is performed using the AdamW optimizer. The best checkpoint is selected based on the accuracy of mechanism classification, as the 8-way prediction task is the most demanding among {\small TSGG} heads and offers a more discriminative selection than $R$@K.

\subsection{Tasks and Metrics}

\textbf{(I) } Our {\small KGE} module is evaluated on three tasks: subject prediction (inferring the head entity given the relation and tail), object prediction (inferring the tail entity given the head and relation), and triplet prediction (assessing the complete triplet). Following \cite{StarE}, we use hits ($H$@K) to measure the percentage of correct entities in top K results and Mean Reciprocal Rank ({\small $MRR$@K}) to evaluate the ranking position of correct entities.

\textbf{(II)} {\small HP} measures the importance of each traffic entity for the driving safety of the ego vehicle \cite{CFHP}. Here, we evaluate how our model prioritizes ego-path relevant entities. We use {\small mAP@K} to assess overall hazards prioritization, {\small MRR@K} to evaluate the rank of the first risky hazard, and Normalized Discounted Cumulative Gain ({\small NDCG@K}) to check ranking quality normalized by ideal order (K=3, 5, 10).

\textbf{(III) }
Graph generation inherently involves three subtasks \cite{xu2017scenegraph}. Predicate Classification ({\small PredCls}) predicts relationships given object categories and locations, which consumes GT segments as inputs incompatible with one-stage designs. Scene Graph Classification ({\small SGCls}) predicts object categories and relations given locations, inapplicable to panoptic-segmentation-based models. Hence, here we only report results on Scene Graph Detection ({\small SGDet}) that jointly predict all components of images, while {\small PredCls}-compatible models \cite{motifs, vctree} are included in comparison IV. Following \cite{yang2022psg}, since $R@\text{K}$ is dominated by frequent relations, we also report $mR@\text{K}$ to evaluate performance uniformly across all relation classes ($\text{K}=\{20,50,100\}$). For two-stage baselines \cite{motifs, vctree}, we pretrain their segmentation branch on Cityscapes and freeze it during the relation training stage.

\textbf{(IV)} We also evaluate how models identify relevant entities and prominent hazards. {\small SGG} methods \cite{motifs, vctree} serve as a baseline due to the lack of dedicated classifiers, using structured predicates to distinguish relevant entities from irrelevant ones, and to derive hazards severity scores from the level of imminent, caution, and info. The classic metrics precision ($P$), recall ($R$), $\text{F}_1$, Area Under the Curve ({\small AUC}) and Binary Mean Absolute Error ({\small B-MAE}) are employed.

\textbf{(V)} To investigate the contribution of each architectural component in {\small HATS}, we conduct a systematic ablation study by progressively removing  the {\small ERES} module, {\small KGE} priors, and depth cues from the full model. We report the relation retrieval performance of each variant on severity, side, and mechanism using $R@\text{K}$ and $mR@\text{K}$ ($\text{K}=\{1,2,3\}$).

\subsection{Results and Discussions}

\begin{table}[!t]
\centering
\caption{Comparison on Three Subtasks of KGE (\%)}
\label{tab:KGE_comparison}
\setlength{\tabcolsep}{3pt}  
\scriptsize
\scalebox{0.95}{
\begin{tabular}{lccccccccc}
\toprule
& \multicolumn{3}{c}{\textbf{Object Prediction}} & \multicolumn{3}{c}{\textbf{Subject Prediction}} & \multicolumn{3}{c}{\textbf{Triplet Prediction}} \\[-1pt] 
\cmidrule(lr){2-4} \cmidrule(lr){5-7} \cmidrule(lr){8-10}
\textbf{Method} & MRR & H@1 & H@10 & MRR & H@1 & H@10 & MRR & H@1 & H@10 \\[-1pt] 
\midrule
StareE \cite{StarE} & 38.66 & 31.50 & 52.63 & 77.21 & 67.97 & 93.36 & 57.94 & 49.73 & 73.00 \\
OUR-1HOP & 88.04 & 79.81 & \textbf{99.77} & \textbf{96.76} & \textbf{94.38} & 99.74 & \textbf{92.40} & \textbf{87.09} & 99.75 \\
OUR-3HOP & \textbf{88.31} & \textbf{80.54} & \textbf{99.77} & 95.96 & 92.77 & \textbf{99.80} & 92.14 & 86.66 & \textbf{99.79} \\[-1pt] 
\bottomrule
\end{tabular}
}
\end{table}
\textbf{(I) } Table~\ref{tab:KGE_comparison} reports the performance of our {\small KGE} module when using 1-hop and 3-hop message passing, compared to the baseline model \cite{StarE}. The consistent gains yielded by our module across all three tasks validate our three key designs. First, literal-aware node initialization jointly encodes categorical and numeric node properties, enabling richer representation of traffic entity characteristics. Second, {\small FiLM} is used to adaptively scale and shift messages via learned parameters derived from the qualifier vector, giving the model more expressive and dynamic control over how edge properties influence message propagation. Third, the transformer-based triplet scorer allows attention-driven interaction between the head entity and the relation before scoring against tail candidates. Additionally, we explicitly maintain separate vocabularies for the relation, supporting heterogeneous qualifier schemas across relations. The substantial performance, with $H$@10 scores near 99.8\%, confirms that our {\small KGE} module captures rich and complete literal and structural patterns from the {\small KG}, validating its reliability as a prior knowledge source for scene understanding.

\begin{table}[!t]
\centering
\caption{Comparison on Hazards Prioritization (\%)}
\label{tab:HP_comparison}
\setlength{\tabcolsep}{3.5pt}  
\scriptsize
\scalebox{0.95}{
\begin{tabular}{lccccccccc}
\toprule
& \multicolumn{3}{c}{\textbf{MAP@K}} 
& \multicolumn{3}{c}{\textbf{MRR@K}} 
& \multicolumn{3}{c}{\textbf{NDCG@K}} \\[-1pt] 
\cmidrule(lr){2-4} \cmidrule(lr){5-7} \cmidrule(lr){8-10}
\textbf{Method} & K=3 & K=5 & K=10 & K=3 & K=5 & K=10 & K=3 & K=5 & K=10 \\[-1pt] 
\midrule
CFHP \cite{CFHP} & 31.66 & 45.64 & 60.97 & 58.41 & 73.90 & 77.78 & 37.94 & 53.40 & 56.95 \\
OUR & \textbf{54.87} & \textbf{68.04} & \textbf{82.90} & \textbf{78.33} & \textbf{88.34} & \textbf{96.60} & \textbf{64.19} & \textbf{77.48} & \textbf{89.47} \\[-1pt] 
\bottomrule
\end{tabular}
}
\vspace{-3mm}
\end{table}
\textbf{(II)} 
Table~\ref{tab:HP_comparison} presents the evaluation results on classic {\small HP} metrics at crucial cutoff values. Our {\small HATS} consistently outperforms {\small CFHP} \cite{CFHP} across all metrics, with especially large gains at lower K values. It indicates that our model ranks the most critical hazards much more accurately and consistently at the top positions, allowing immediate driver alert in time-critical situations. This stems from several novel designs working in concert. First, the {\small ERES} module establishes a strong foundation by learning to attend to path-relevant entities through cross-attention between the ego path and traffic entities. The discrimination is further refined by the {\small TSGG} heads during joint training. Moreover, the {\small KGE} representation in the pair descriptor, grounded in real-life traffic accident data, encodes structured domain knowledge about hazard severity and consequence patterns that pure visual methods lack. The prior aggregator further reinforces this by attending over severity-related {\small KG} node groups, injecting accident priors directly into the severity head. 

\textbf{(III)} 
The {\small SGCls} results in Table~\ref{tab:sgg_comparison} validate the challenges identified in Sec. \ref{section:related_work} regarding the application of {\small SG} and {\small PSG} in traffic scenarios. Since majority traffic entities are irrelevant to the main driving maneuver with only a few critical to attend, the baseline models favor frequent predicates at the expense of the tail categories, resulting in low $mR@\text{K}$. By contrast, our {\small ERES} and {\small TSGG} modules decompose the prediction into well-defined subtasks toward the unified objective. The {\small ERES} module explicitly filters the dominant         `irrelevant' predicate from the relation prediction space, rebalancing the distribution toward meaningful hazard-aware predicates. The pair descriptor that integrates visual, geometric, semantic, and knowledge-based representations encourages {\small TSGG} heads to reason from multiple complementary aspects, rather than being blinded by superficial frequency. The gains in both $R@\text{K}$ and $mR@\text{K}$ demonstrate that {\small HATS} strikes a balance between maximizing the retrieval of correct relations in the surrounding environment and maintaining a consistent prediction quality of less frequent but safety-critical interactions.

\begin{table}[!t]
\centering
\caption{Comparison on {\small SGDet} Task (\%)}
\label{tab:sgg_comparison}
\renewcommand{\arraystretch}{0.9}
\scriptsize
\scalebox{0.95}{
\begin{tabular}{lccc}
\toprule
\textbf{Method} & \textbf{R@20 / mR@20} & \textbf{R@30 / mR@30} & \textbf{R@50 / mR@50} \\[-2pt]
\midrule
MOTIFS \cite{motifs} & 25.99 / 7.39 & 29.65 / 9.37 & 29.68 / 9.38 \\
VCTREE \cite{vctree} & 26.95 / 4.12 & 29.60 / 5.23 & 29.60 / 5.23 \\
PSGTR \cite{yang2022psg} & 17.95 / 5.53 & 23.71 / 12.20 & 23.72 / 12.21 \\
OUR    & \textbf{30.47} / \textbf{53.75} & \textbf{50.00} / \textbf{67.63} & \textbf{62.79} / \textbf{76.13} \\[-2pt] 
\bottomrule
\end{tabular}
}
\vspace{-3mm}
\end{table}
\textbf{(IV)} Table~\ref{tab:classification_comparison} reflects the precision-recall imbalance of baselines in relevance classification, suggesting that their conservative predicate predictions struggle to identify all relevant entities, likely due to their lack of path-awareness. Our {\small ERES} module directly addresses this by modeling the interaction between the ego path and each entity through learnable cross-attention. The fused feature \eqref{eq:fused_feature} that combines entity content, global context summary, and a gated path-conditioned signal provides a richer and more targeted representation for relevance classification. The strong relevance performance is further attributed to the joint training after warm-up. The exclusive operation in the {\small TSGG} module on the filtered candidates and the explicit leverage of relevance logits in constructing semantic representation fed into all {\small TSGG} heads allow task-specific gradients to flow back into the {\small ERES} module for continuous refinement. In prominence classification, {\small HATS} still outperforms specialized competitor \cite{CFHP} and visual-driven baselines, attributable to the multi-source pair descriptor and accident-informed severity head.

\begin{table}[!t]
\centering
\caption{Comparison on Two Classification Tasks (\%)}
\label{tab:classification_comparison}
\setlength{\tabcolsep}{3pt}
\renewcommand{\arraystretch}{0.9}
\scriptsize
\scalebox{0.95}{
\begin{tabular}{clccccc}
\toprule
\textbf{Task} & \textbf{Method} & \textbf{P}(\%) & \textbf{R}(\%) & $\textbf{F}_1$(\%) & \textbf{AUC}(\%) & \textbf{MAE}$\downarrow$ \\[-1pt]
\midrule
\multirow{6}{*}{\shortstack{\textbf{Entity} \\ \textbf{Relevance}}}
& MOTIFS-predcls \cite{motifs} & 82.96 & 43.46 & 53.42 & 81.14 & 0.323  \\
& MOTIFS-sgdet \cite{motifs} & 85.73 & 50.90 & 61.61 & 77.11 & 0.270  \\
& VCTREE-predcls \cite{vctree} & 83.59 & 43.40 & 53.05 & 81.63 & 0.315  \\
& VCTREE-sgdet \cite{vctree} & 88.15 & 36.87 & 49.84 & 76.19 & 0.319  \\
& PSGTR \cite{yang2022psg} & 87.62 & 31.74 & 44.24 & 62.88 & 0.327 \\
& OUR   & \textbf{96.96} & \textbf{94.07} & \textbf{95.03} & \textbf{97.20} & \textbf{0.047} \\[-2pt] 
\midrule
\multirow{7}{*}{\shortstack{\textbf{Entity} \\ \textbf{Prominence}}}
& MOTIFS-predcls \cite{motifs} & 85.10 & 66.68 & 72.51 & 76.61 & 0.370  \\
& MOTIFS-sgdet \cite{motifs} & 86.25 & 61.33 & 69.70 & 72.56 & 0.367  \\
& VCTREE-predcls \cite{vctree} & 83.70 & 66.53 & 71.90 & 75.33 & 0.378 \\
& VCTREE-sgdet \cite{vctree} & 85.44 & 58.27 & 67.23 & 69.75 & 0.416 \\
& PSGTR \cite{yang2022psg} & 87.53 & 38.32 & 50.92 & 62.20 & 0.477 \\
& CFHP \cite{CFHP} & 88.14 & 49.77 & 62.49 & 62.39 & 0.233 \\
& OUR    & \textbf{95.85} & \textbf{87.78} & \textbf{90.89} & \textbf{92.17} & \textbf{0.103} \\[-2pt] 
\bottomrule
\end{tabular}
}
\end{table}

\begin{table}[!t]
\centering
\caption{Ablation Study of HATS components}
\label{tab:relation_comparison}
\setlength{\tabcolsep}{3pt}
\renewcommand{\arraystretch}{0.9}
\scriptsize
\scalebox{0.95}{
\begin{tabular}{llccc}
\toprule
\textbf{Relation} & \textbf{Method} & \textbf{R@1/mR@1} & \textbf{R@2/mR@2} & \textbf{R@3/mR@3} \\[-1pt]
\midrule
\multirow{6}{*}{\textbf{Severity}} 
& HATS {\tiny w/o ERES} & 31.27 / 35.32	& 44.42 / 50.05	& 48.42 / 54.89 \\
& HATS {\tiny w/o KGE} & 42.03	/ 44.99 & 60.00 / 63.02 &	65.39 / 68.89 \\
& HATS {\tiny w/o Depth} & 41.21 / 43.15 & 60.47 / 61.69 & 65.02 / 66.96 \\
& HATS {\tiny w/o KGE+Depth} & 39.46 / 43.25 &	58.39 / 61.09 & 63.61 / 66.43 \\
& HATS {\tiny w/o ERES+KGE+Depth} & 26.29 / 31.68 &	42.01 / 47.91 &	47.56 / 53.39 \\
& HATS    & \textbf{80.10} / \textbf{80.72} & \textbf{92.19} / \textbf{92.76} & \textbf{93.93} / \textbf{94.99} \\[-2pt] 
\midrule
\multirow{6}{*}{\textbf{Side}} 
& HATS {\tiny w/o ERES} & 38.68 / 39.23	& 48.29	/ 49.13 & 49.37 / 50.46 \\
& HATS {\tiny w/o KGE} & 51.33 / 50.54 & 65.11 / 65.00	& 66.70 / 66.87 \\
& HATS {\tiny w/o Depth} & 51.20 / 51.22 &	65.31 / 65.23	& 66.34 / 66.41 \\
& HATS {\tiny w/o KGE+Depth} & 49.65 / 50.39 & 63.33 / 63.30 & 64.80 / 64.99 \\
& HATS {\tiny w/o ERES+KGE+Depth} & 37.40 / 37.62 &	47.31 / 47.21 & 48.63 / 48.76 \\
& HATS    & \textbf{82.40} / \textbf{81.18} & \textbf{93.73} / \textbf{93.06} & \textbf{94.53} / \textbf{94.01} \\[-2pt] 
\midrule
\multirow{6}{*}{\textbf{Mechanism}} 
& HATS {\tiny w/o ERES} & 30.33 / 32.32 & 39.14 / 42.45	& 42.46 / 45.58 \\
& HATS {\tiny w/o KGE} & 42.52 / 43.62 & 57.27 / 59.85 & 62.78 / 65.21 \\
& HATS {\tiny w/o Depth} & 35.65 / 37.59 &	51.97 / 52.82 & 59.23 / 59.51 \\
& HATS {\tiny w/o KGE+Depth} & 36.47 / 37.97 &	52.12 / 54.33 & 58.98 / 61.26 \\
& HATS {\tiny w/o ERES+KGE+Depth} & 30.55 / 34.62 & 40.48 / 45.61 & 44.38 / 49.52 \\
& HATS    & \textbf{73.16} / \textbf{73.65} & \textbf{84.01} / \textbf{85.91} & \textbf{88.62} / \textbf{90.37}  \\[-2pt] 
\bottomrule
\end{tabular}
}
\end{table} 

\begin{figure}[!t]
\centering
\setlength{\tabcolsep}{1pt}
\begin{tabular}{ccc}
\includegraphics[width=0.32\columnwidth]{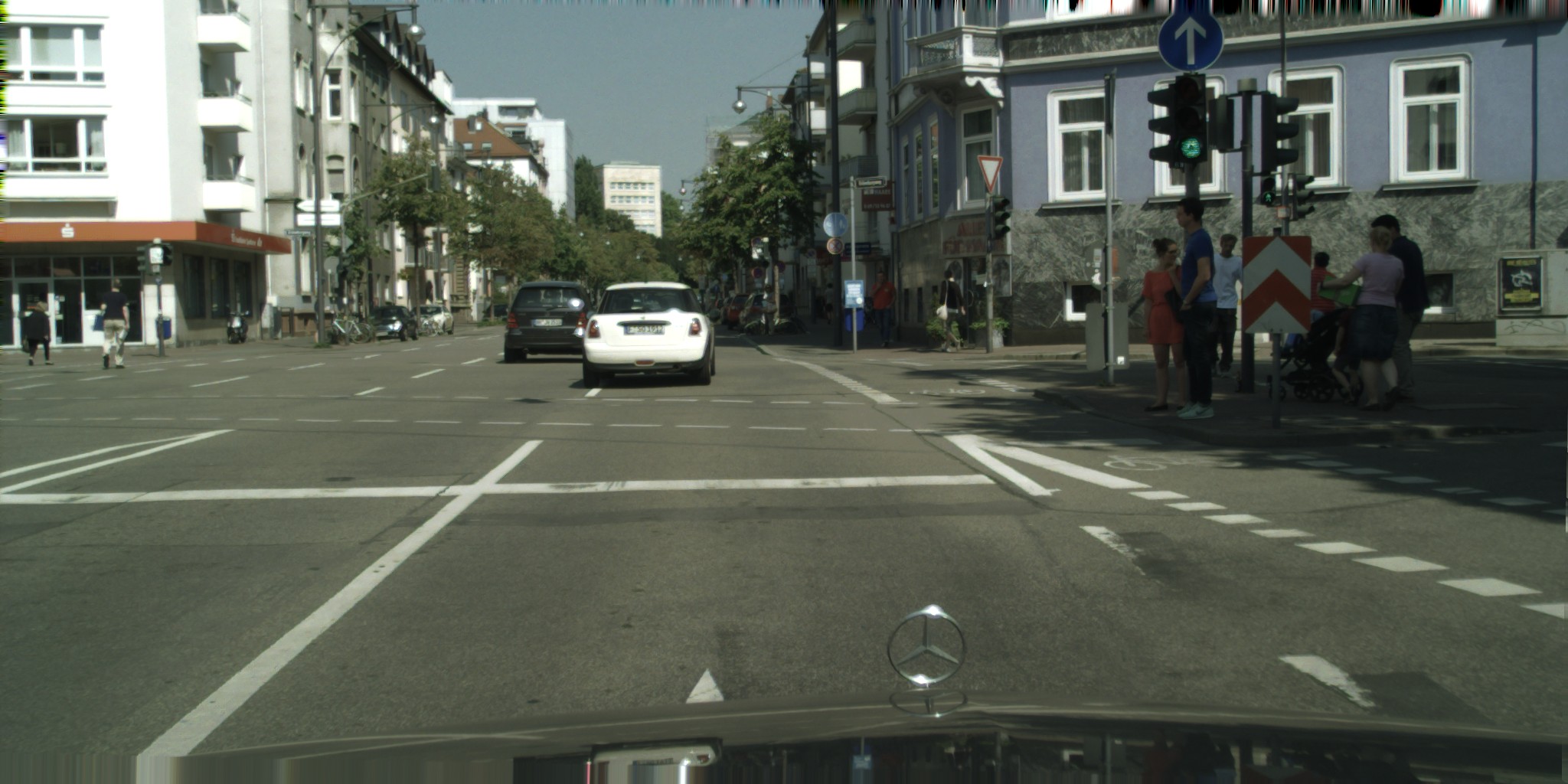} &
\includegraphics[width=0.32\columnwidth]{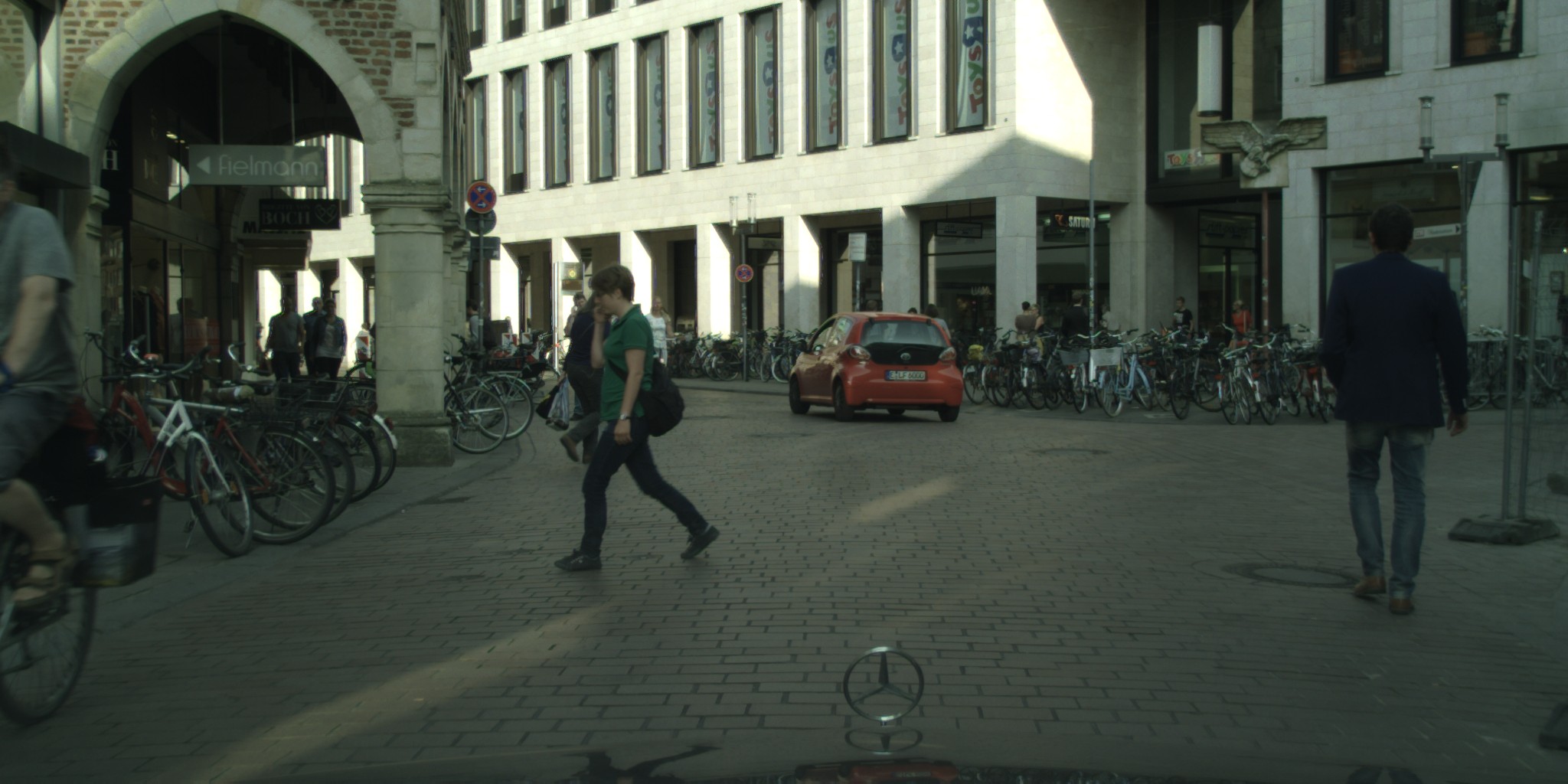} &
\includegraphics[width=0.32\columnwidth]{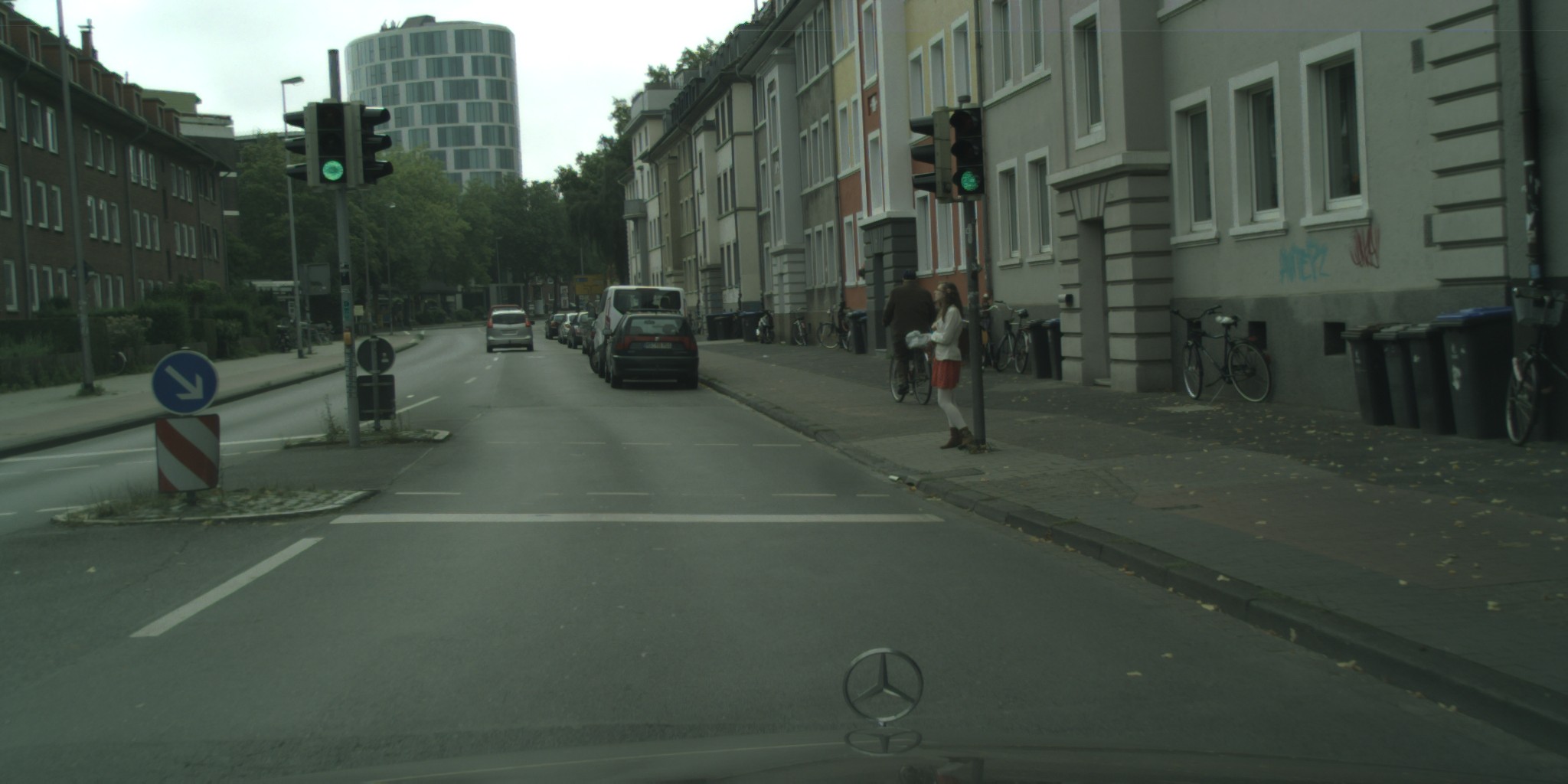} \\[-2pt]
\includegraphics[width=0.32\columnwidth]{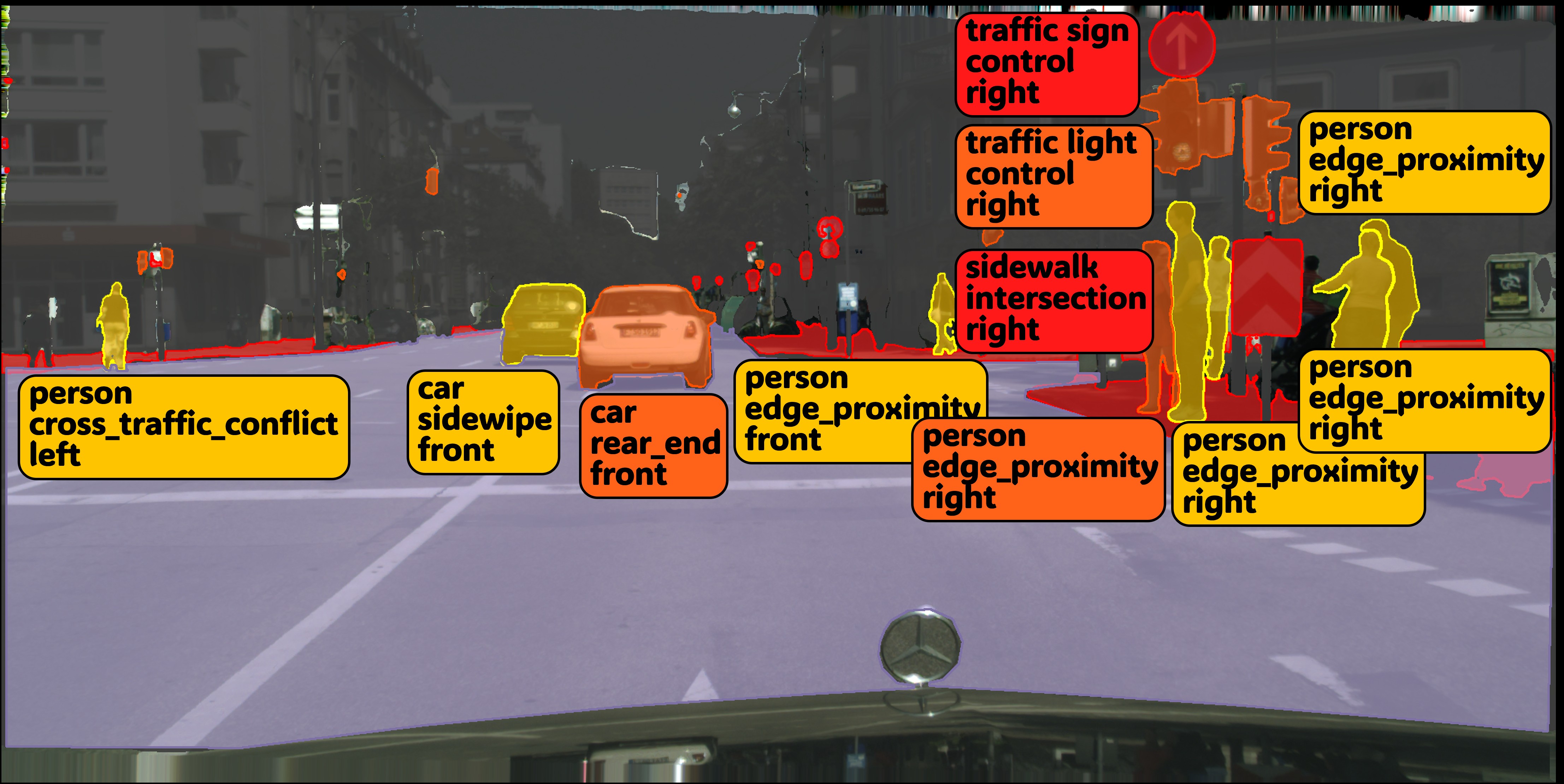} &
\includegraphics[width=0.32\columnwidth]{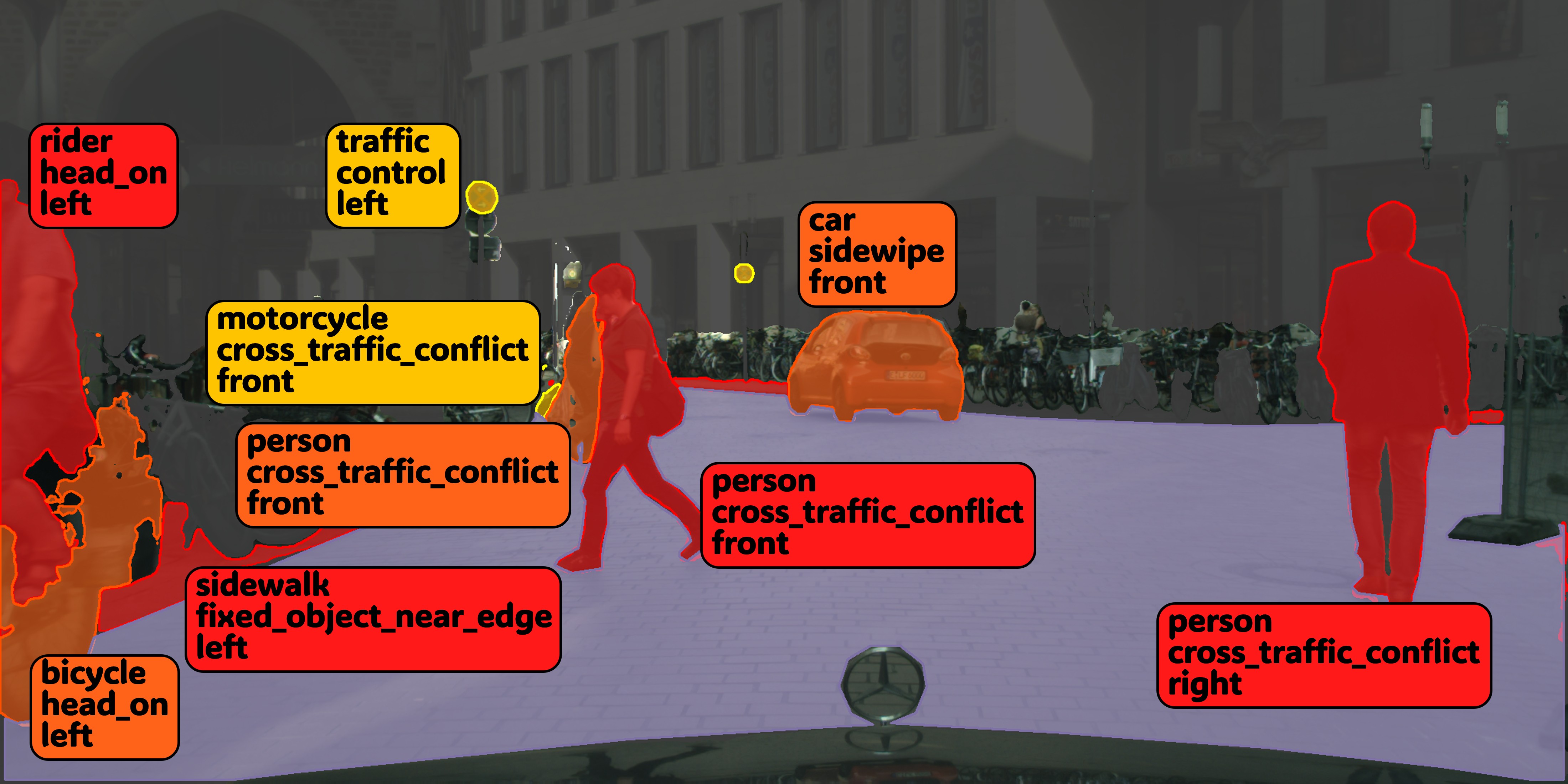} &
\includegraphics[width=0.32\columnwidth]{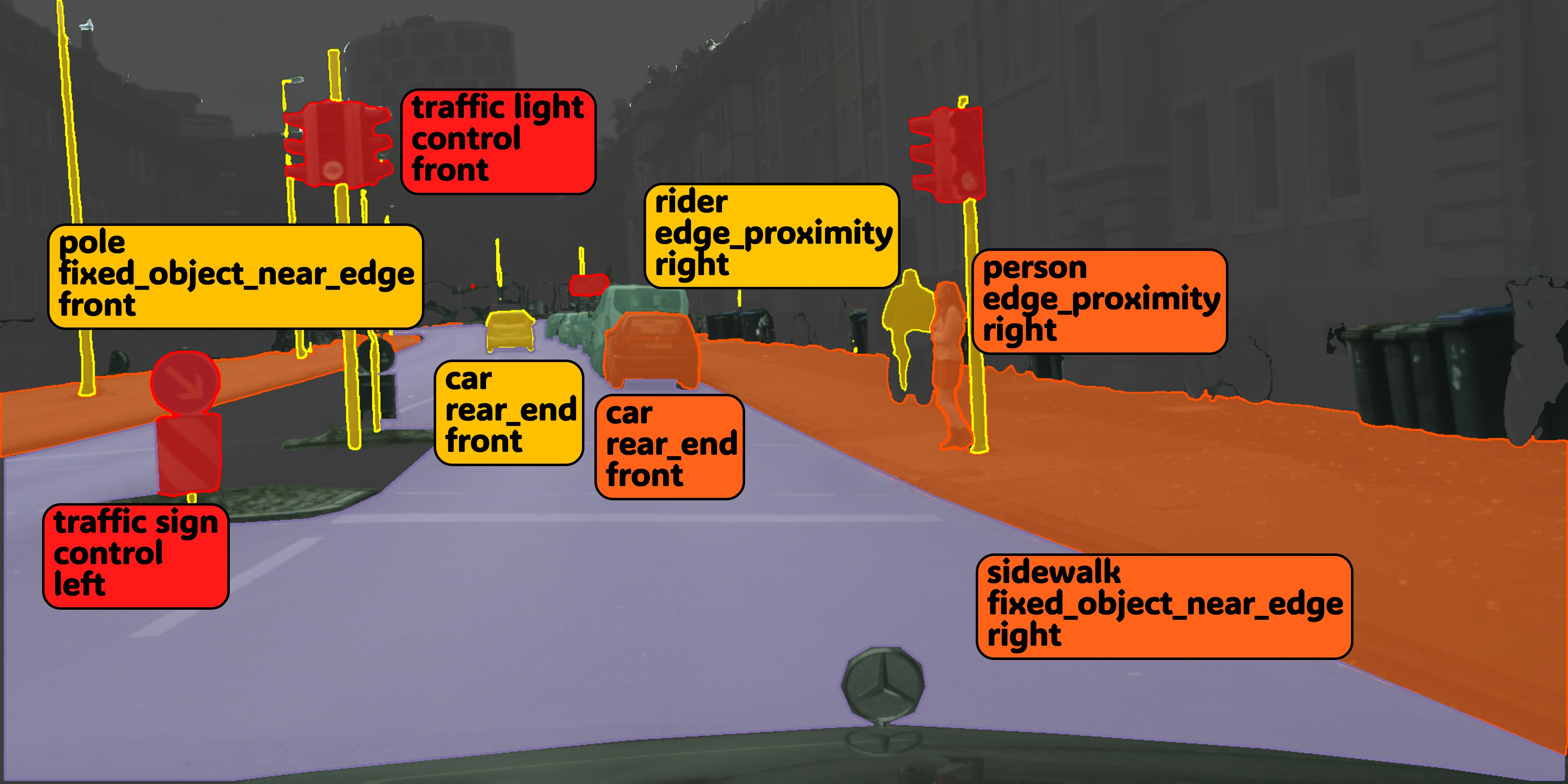} \\[-2pt]
\multicolumn{3}{c}{\includegraphics[width=0.97\columnwidth]{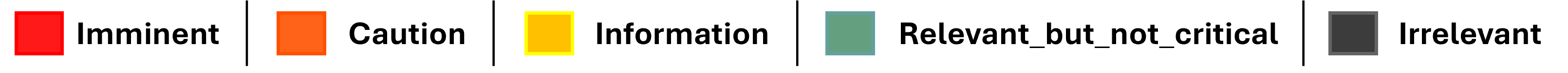}} \\[-6pt]
\end{tabular}
\caption{Our ego-centric hazard-aware TSGs of traffic images}
\label{fig:viz_TSG}
\vspace{-5mm}
\end{figure}

\textbf{(V)} Table~\ref{tab:relation_comparison} highlights the critical role of the {\small ERES} module: removing it alone results in more than a 40\% drop in $R$@1 across all three tasks. The nearly identical performance of {\small HATS} {\footnotesize w/o ERES+KGE+Depth} and {\small HATS} {\footnotesize w/o ERES} further indicates that {\small ERES} is a prerequisite for {\small KGE} and depth cues to be effective. Without {\small ERES}, all panoptic queries, including irrelevant entities such as sky and distant parked cars are passed to {\small TSGG} heads with uniform weights, overwhelming relation prediction with unstructured context and hindering the learning of discriminative representations, regardless of feature richness.
{\small HATS} {\footnotesize w/o KGE} results suggest the significance of the structured {\small KGE} priors in separating geometrically or visually similar but semantically distinct scenarios, such as an approaching vehicle from opposite lane implying a serious head-on crash. Without {\small KGE}, {\small TSGG} heads must search for clues from the high-dimensional but unstructured appearance features alone, which is especially unrecoverable for severity prediction. 
Depth features explicitly encode relative distance and lateral offset of each entity that could directly impact severity and effect mechanism, for instance, a sidewipe from the adjacent car versus irrelevance from a parking car over the barrier. Without disparity-derived cues, pair descriptor will force {\small TSGG} heads to rely on {\small RGB} appearance and semantic embeddings, which are insufficient to resolve spatial ambiguity, evidenced by the drops in {\small HATS} {\footnotesize w/o Depth}.
The marginal improvement of {\small HATS} {\footnotesize w/o KGE+Depth} over {\small HATS} {\footnotesize w/o Depth} suggests that {\small KGE} priors, deprived of the depth-derived geometric grounding, introduce conflicting semantic bias that slightly degrades prediction. This feature interference effect further validates our joint design. 
Overall, the huge gap between any single-component ablation and the full model demonstrates that all components are mutually reinforcing and jointly necessary for an optimal capacity.

\section{Conclusions}
In this paper, we present {\small HATS}, an ego-centric hazard-aware TSGG framework that addresses the limitations of existing {\small SGG} and {\small PSG} methods in safety-critical autonomous driving contexts. Its supplementary {\small KG} branch embeds structured real-life traffic accident records through literal-aware node initialization, {\small FiLM}-based qualifier-aware message passing, and a transformer-based triplet scorer. Its main {\small SG} branch consists of three modules. The {\small ERES} module built upon {\small PS} module picks relevant candidates through learnable cross-attention between the ego path and traffic entities for downstream {\small TSGG} module. The dedicated TSGG heads, supported by the diverse-feature-enriched pair descriptor and the knowledge-grounding prior aggregator, decompose relation prediction into structured subtasks, jointly trained to understand the effect mechanism, the relative side, and the severity level with respect to the ego vehicle. Ablation studies confirm the  indispensable and synergistic role of each module in overall performance. Extensive experiments demonstrate the outperformance of {\small HATS} in {\small HP}, relevance classification, prominence classification, {\small SGDet}, and relation predictions. These results validate the effectiveness of integrating structured accident knowledge with ego-centric visual reasoning for comprehensive yet hazard-aware traffic scene understanding. Our future work will extend {\small HATS} to video-based reasoning and stronger segmentation backbones to improve performance in broader driving environments.






\bibliographystyle{ieeetr}
\bibliography{reference}

@book{who_mobile_2011,
  title = {Mobile phone use: A growing problem of driver distraction},
  publisher = {World Health Org.},
  address = {Geneva},
  year = {2011},
  author = {{World Health Org.}}
}

@article{Fagnant,
author = {Fagnant, Daniel and Kockelman, Kara},
year = {2015},
month = {07},
pages = {},
title = {Preparing a nation for autonomous vehicles: Opportunities, barriers and policy recommendations},
volume = {77},
journal = {Transp. Res. Part A: Policy and Pract.},
doi = {10.1016/j.tra.2015.04.003}
}

@article{2022_Choudhary,
author = {P Choudhary and A Gupta and N R Velaga},
title = {Perceived risk vs actual driving performance during distracted driving: A comparative analysis of phone use and other secondary distractions},
journal = {Transp. Res. Part F: Traffic Psychol. and Behav.},
year = {2022},
volume = {86},
publisher = {Elsevier},
month = {apr},
url = {https://doi.org/10.1016/j.trf.2022.03.001},
pages = {296--315},
doi = {10.1016/j.trf.2022.03.001}
}

@ARTICLE{9758639,
  author={Yusof, Nidzamuddin Md. and Karjanto, Juffrizal and Hassan, Muhammad Zahir and Terken, Jacques and Delbressine, Frank and Rauterberg, Matthias},
  journal={IEEE Trans. on Intell. Transp. Syst.}, 
  title={Reading During Fully Automated Driving: A Study of the Effect of Peripheral Vis. and Haptic Inf. on Situation Awareness and Mental Workload}, 
  year={2022},
  volume={23},
  number={10},
  pages={19136-19144},
  doi={10.1109/TITS.2022.3165192}}

@INPROCEEDINGS{PanopticSegmentation,
  author={Kirillov, Alexander and He, Kaiming and Girshick, Ross and Rother, Carsten and Dollár, Piotr},
  booktitle={CVPR}, 
  title={Panoptic Segmentation}, 
  year={2019},
  volume={},
  number={},
  pages={9396-9405},
  doi={10.1109/CVPR.2019.00963}}

@ARTICLE{9423525,
  author={Yu, Shih-Yuan and Malawade, Arnav Vaibhav and Muthirayan, Deepan and Khargonekar, Pramod P. and Faruque, Mohammad Abdullah Al},
  journal={IEEE Trans. on Intell. Transp. Syst.}, 
  title={Scene-Graph Augmented Data-Driven Risk Assessment of Auton. Vehicle Decisions}, 
  year={2022},
  volume={23},
  number={7},
  pages={7941-7951},
  doi={10.1109/TITS.2021.3074854}}

@inproceedings{yang2022psg,
    author = {Yang, Jingkang and Ang, Yi Zhe and Guo, Zujin and Zhou, Kaiyang and Zhang, Wayne and Liu, Ziwei},
    title = {Panoptic Scene Graph Gener.},
    booktitle = {ECCV},
    year = {2022}
}

@article{VisualGenome,
author = {Krishna, Ranjay and Zhu, Yuke and Groth, Oliver and Johnson, Justin and Hata, Kenji and Kravitz, Joshua and Chen, Stephanie and Kalantidis, Yannis and Li, Li-Jia and Shamma, David A. and Bernstein, Michael S. and Fei-Fei, Li},
title = {Vis. Genome: Connecting Lang. and Vision Using Crowdsourced Dense Image Annotations},
year = {2017},
issue_date = {May 2017},
publisher = {Kluwer Academic Publishers},
address = {USA},
volume = {123},
number = {1},
issn = {0920-5691},
url = {https://doi.org/10.1007/s11263-016-0981-7},
doi = {10.1007/s11263-016-0981-7},
journal = {Int. J. Comput. Vision},
month = may,
pages = {32–73},
numpages = {42}
}

@book{NHTSA1,
  title = {NHTSA Field Crash Investigation 2021 Coding and Editing Manual},
  publisher = {Nat. Highway Traffic Saf. Admin.},
  year = {2022},
  author = {Nat. Highway Traffic Saf. Admin.}
}

@book{NHTSA2,
  title = {Crash Investigation Sampling System 2021 Analytical User’s Manual},
  publisher = {Nat. Highway Traffic Saf. Admin.},
  year = {2022},
  author = {Radja, Gregory A and Noh, Eun-Young and Zhang, Fan}
}

@INPROCEEDINGS{SGG,
  author={Johnson, Justin and Krishna, Ranjay and Stark, Michael and Li, Li-Jia and Shamma, David A. and Bernstein, Michael S. and Fei-Fei, Li},
  booktitle={CVPR}, 
  title={Image retrieval using scene graphs}, 
  year={2015},
  volume={},
  number={},
  pages={3668-3678},
  doi={10.1109/CVPR.2015.7298990}}

@inproceedings{xu2017scenegraph,
  title={Scene Graph Gener. by Iterative Message Passing},
  author={Xu, Danfei and Zhu, Yuke and Choy, Christopher and Fei-Fei, Li},
  booktitle={CVPR},
  year={2017}
 }

@inproceedings{lu2016visual,
  title={Vis. Relationship Detection with Lang. Priors},
  author={Lu, Cewu and Krishna, Ranjay and Bernstein, Michael and Fei-Fei, Li},
  booktitle={Eur. Conf. on Comput. Vision},
  year={2016}
}

@INPROCEEDINGS{8954048,
  author={Chen, Tianshui and Yu, Weihao and Chen, Riquan and Lin, Liang},
  booktitle={CVPR}, 
  title={Knowl.-Embedded Routing Network for Scene Graph Gener.}, 
  year={2019},
  volume={},
  number={},
  pages={6156-6164},
  doi={10.1109/CVPR.2019.00632}}

@article{Reltr,
  title={Reltr: Relation transformer for scene graph generation},
  author={Cong, Yuren and Yang, Michael Ying and Rosenhahn, Bodo},
  journal={IEEE Trans. on Pattern Anal. and Mach. Intell.},
  year={2023},
  publisher={IEEE}
}

@INPROCEEDINGS{9575491,
  author={Tian, Yafu and Carballo, Alexander and Li, Ruifeng and Takeda, Kazuya},
  booktitle={IV}, 
  title={RSG-Net: Towards Rich Sematic Relationship Prediction for Intell. Vehicle in Complex Environments}, 
  year={2021},
  volume={},
  number={},
  pages={546-552},
  doi={10.1109/IV48863.2021.9575491}}

@INPROCEEDINGS{9197057,
  author={Li, Chengxi and Meng, Yue and Chan, Stanley H. and Chen, Yi-Ting},
  booktitle={ICRA}, 
  title={Learning 3D-aware Egocentric Spatial-Temporal Interaction via Graph Convolutional Networks}, 
  year={2020},
  volume={},
  number={},
  pages={8418-8424},
  doi={10.1109/ICRA40945.2020.9197057}}

@INPROCEEDINGS{RS10K,
  author={Guo, Yunfei and Yin, Fei and Li, Xiao-Hui and Yan, Xudong and Xue, Tao and Mei, Shuqi and Liu, Cheng-Lin},
  booktitle={2023 IEEE/CVF Int. Conf. on Comput. Vision (ICCV)}, 
  title={Vis. Traffic Knowl. Graph Gener. from Scene Images}, 
  year={2023},
  volume={},
  number={},
  pages={21547-21556},
  doi={10.1109/ICCV51070.2023.01975}}

@ARTICLE{HDD,
  author={Zhou, Yuchen and Zhang, Yue and Zhao, Zhanwei and Zhang, Kaidong and Gou, Chao},
  journal={IEEE J. of Radio Freq. Identification}, 
  title={Toward Driving Scene Understanding: A Paradigm and Benchmark Dataset for Ego-Centric Traffic Scene Graph Representation}, 
  year={2022},
  volume={6},
  number={},
  pages={962-967},
  doi={10.1109/JRFID.2022.3207017}}

@inproceedings{Cityscapes,
title={The Cityscapes Dataset for Semantic Urban Scene Understanding},
author={Cordts, Marius and Omran, Mohamed and Ramos, Sebastian and Rehfeld, Timo and Enzweiler, Markus and Benenson, Rodrigo and Franke, Uwe and Roth, Stefan and Schiele, Bernt},
booktitle={CVPR},
year={2016}
}

@article{WORDNET,
author = {Miller, George A.},
title = {WordNet: a lexical database for English},
year = {1995},
issue_date = {Nov. 1995},
publisher = {Assoc. for Comput. Machinery},
address = {New York, NY, USA},
volume = {38},
number = {11},
issn = {0001-0782},
url = {https://doi.org/10.1145/219717.219748},
doi = {10.1145/219717.219748},
journal = {Commun. ACM},
month = nov,
pages = {39–41},
numpages = {3}
}

@inproceedings{ConceptNet,
author = {Speer, Robyn and Chin, Joshua and Havasi, Catherine},
title = {ConceptNet 5.5: an open multilingual graph of general knowledge},
year = {2017},
booktitle = {AAAI},
pages = {4444–4451},
numpages = {8},
location = {San Francisco, California, USA},
}

@inproceedings{you2020CTA,
    title     = "{Traffic Accident Benchmark for Causality Recognit.}",
    author    = {You, Tackgeun and Han, Bohyung},
    booktitle = {ECCV},
    year      = {2020}
}

@INPROCEEDINGS{8578469,
  author={Suzuki, Tomoyuki and Kataoka, Hirokatsu and Aoki, Yoshimitsu and Satoh, Yutaka},
  booktitle={CVPR}, 
  title={Anticipating Traffic Accidents with Adaptive Loss and Large-Scale Incident DB}, 
  year={2018},
  volume={},
  number={},
  pages={3521-3529},
  doi={10.1109/CVPR.2018.00371}}

@ARTICLE{9424477,
  author={Lin, Da-Jie and Chen, Mu-Yen and Chiang, Hsiu-Sen and Sharma, Pradip Kumar},
  journal={IEEE Trans. on Intell. Transp. Syst.}, 
  title={Intell. Traffic Accident Prediction Model for Internet of Vehicles With Deep Learning Approach}, 
  year={2022},
  volume={23},
  number={3},
  pages={2340-2349},
  doi={10.1109/TITS.2021.3074987}}

@INPROCEEDINGS{9723740,
  author={Salam, Sayeed and Islam, Md Shihabul and Ahmed, Fawaz and Khan, Latifur and Kim, Dohyeong and Allo, Nicholas and Nwariaku, Ohwofiemu},
  booktitle={2021 IEEE 4th Int. Conf. on Artif. Intell. and Knowl. Eng. (AIKE)}, 
  title={Exploring the roles of social media data to identify the locations and severity of road traffic accidents}, 
  year={2021},
  volume={},
  number={},
  pages={62-71},
  doi={10.1109/AIKE52691.2021.00016}}

@ARTICLE{s2tld,
author={Yang, Xue and Yan, Junchi and Liao, Wenlong and Yang, Xiaokang and Tang, Jin and He, Tao},
journal={ IEEE Trans. on Pattern Anal. \& Mach. Intell. },
title={{ SCRDet++: Detecting Small, Cluttered and Rotated Objects via Instance-Level Feature Denoising and Rotation Loss Smoothing }},
year={2023},
volume={45},
number={02},
ISSN={1939-3539},
pages={2384-2399},
doi={10.1109/TPAMI.2022.3166956},
url = {https://doi.ieeecomputersociety.org/10.1109/TPAMI.2022.3166956},
publisher={IEEE Comput. Soc.},
address={Los Alamitos, CA, USA},
month=feb}

@inproceedings{mask2former,
  title={Masked-attention Mask Transformer for Universal Image Segmentation},
  author={Bowen Cheng and Ishan Misra and Alexander G. Schwing and Alexander Kirillov and Rohit Girdhar},
  booktitle = {CVPR},
  year={2022}
}

@inproceedings{TransR,
author = {Lin, Yankai and Liu, Zhiyuan and Sun, Maosong and Liu, Yang and Zhu, Xuan},
title = {Learning entity and relation embeddings for knowledge graph completion},
year = {2015},
isbn = {0262511290},
booktitle = {AAAI},
pages = {2181–2187},
numpages = {7},
location = {Austin, Texas},
}

@inproceedings{StarE,
    title = "Message Passing for Hyper-Relational Knowl. Graphs",
    author = "Galkin, Mikhail and Trivedi, Priyansh  and Maheshwari, Gaurav  and Usbeck, Ricardo  and Lehmann, Jens",
    booktitle = "Proc. of the 2020 Conf. on Empirical Methods in Natural Lang. Process.",
    month = nov,
    year = "2020",
    address = "Online",
    publisher = "Assoc. for Comput. Linguistics",
    url = "https://aclanthology.org/2020.emnlp-main.596/",
    doi = "10.18653/v1/2020.emnlp-main.596",
    pages = "7346--7359",
}

@inproceedings{FiLM,
author = {Perez, Ethan and Strub, Florian and de Vries, Harm and Dumoulin, Vincent and Courville, Aaron},
title = {FiLM: visual reasoning with a general conditioning layer},
year = {2018},
isbn = {978-1-57735-800-8},
booktitle={AAAI},
articleno = {483},
numpages = {10},
location = {New Orleans, Louisiana, USA},
}

@inproceedings{vctree,
  title={Learning to Compose Dynamic Tree Structures for Vis. Contexts},
  author={Tang, Kaihua and Zhang, Hanwang and Wu, Baoyuan and Luo, Wenhan and Liu, Wei},
  booktitle= "CVPR",
  year={2019}
}

@ARTICLE{CFHP,
  author={Huang, Yaoqi and Wang, Xiuying},
  journal={IEEE Trans. on Intell. Transp. Syst.}, 
  title={Hazards Prioritization With Cogn. Attention Maps for Supporting Driving Decision-Making}, 
  year={2024},
  volume={25},
  number={11},
  pages={16221-16234},
  doi={10.1109/TITS.2024.3413675}}

@INPROCEEDINGS {motifs,
author = {Zellers, Rowan and Yatskar, Mark and Thomson, Sam and Choi, Yejin },
booktitle = {CVPR},
title = {{Neural Motifs: Scene Graph Parsing with Global Context }},
year = {2018},
volume = {},
ISSN = {},
pages = {5831-5840},
doi = {10.1109/CVPR.2018.00611},
url = {https://doi.ieeecomputersociety.org/10.1109/CVPR.2018.00611}}

\end{document}